\documentclass{ntupaper} % DO NOT CHANGE THIS
\usepackage[]{url}  % DO NOT CHANGE THIS
\usepackage{graphicx} % DO NOT CHANGE THIS
\usepackage{wrapfig} % Added for wrapfigure environment
\usepackage{natbib}  % DO NOT CHANGE THIS AND DO NOT ADD ANY OPTIONS TO IT
\usepackage{caption} % DO NOT CHANGE THIS AND DO NOT ADD ANY OPTIONS TO IT
\usepackage{algorithm}
\usepackage{algorithmic}
\usepackage{xparse}
\usepackage{longtable} % Added for longtable environment
\usepackage{soul}
\usepackage{listings}
\usepackage{svg}
\lstdefinelanguage{json}{
  morestring=[b]",
  morecomment=[l]{//},
  morekeywords={true,false,null},
  sensitive=false,
}
\usepackage{newfloat}
\usepackage{listings}
\DeclareCaptionStyle{ruled}{labelfont=normalfont,labelsep=colon,strut=off} % DO NOT CHANGE THIS
\floatstyle{ruled}
\newfloat{listing}{tb}{lst}{}
\floatname{listing}{Listing}

\usepackage{booktabs}
\usepackage{multirow}
\usepackage{array}

\usepackage{adjustbox}

\newcommand{\hist}{\mathcal{H}}
\newcommand{\methodname}{MNIST-PRO}
\newcommand{\method}{\texttt{\methodname}}

\newcommand{\geminipro}{\textsc{Gemini-3.1-Pro-Preview}}
\newcommand{\geminisix}{\textsc{Gemini-3.6-Flash}}
\newcommand{\geminiseven}{\textsc{Gemini-3.7-Flash}}
\newcommand{\claudesonnet}{\textsc{Claude-5-Sonnet}}
\newcommand{\claudeopus}{\textsc{Claude-5-Opus}}
\newcommand{\claudefable}{\textsc{Claude-5-Fable}}
\newcommand{\gptterra}{\textsc{GPT-5.6-Terra}}
\newcommand{\gptsol}{\textsc{GPT-5.6-Sol}}
\newcommand{\gptsolxhigh}{\textsc{GPT-5.6-Sol} (xhigh)}
\newcommand{\qwen}{\textsc{Qwen-3.8-27B}}
\newcommand{\glm}{\textsc{GLM-4.6V}}

\usepackage{amsmath}
\usepackage{cleveref}
\usepackage{amssymb}
\usepackage[table]{xcolor}

\title{MNIST-PRO: MNIST is Back as a Partially Observable World for AI Agents}

\addaffiliation{lab}{DeCLaRe Lab, Nanyang Technological University, Singapore}

\addaffiliation{astar}{Agency for Science, Technology, and Research (A*STAR), Singapore}

\author[lab,astar]{Vernon Toh}
\author[lab]{Navonil Majumder}
\author[astar]{Zhengyuan Liu}
\author[astar]{Nancy F. Chen}
\author[lab]{Soujanya Poria}             % keys and literal markers mix freely
\correspondingauthor{Soujanya Poria}{soujanya.poria@ntu.edu.sg}
\declareorg{declare-lab}
\code{\methodname}
\exportmetadata[2026]{doe2026ntupaper}  % writes a BibTeX + arXiv sidecar

\abstract{
AI agents in partially observable environments need to coordinate active sensing with working memory to maintain an evolving perceptual state. 
However, existing benchmarks struggle to isolate this perceptual-state construction and interpretation capability because they introduce physical and control complexities. 
We address this with \method{}, a benchmark that isolates agentic perception by converting MNIST digit recognition into a sequential, glimpse-based search task with lookback constraints. 
We evaluate ten multimodal models across four memory representations, including raw visual history, textual states, structured metric grid maps, and a consolidated visual canvas. 
While models excel under full observability, partial observability exposes a clear performance gap. 
We identify three distinct bottlenecks. 
First, perceptual-state construction and interpretation present a challenge, as agents struggle to integrate fragmented glimpses. 
Second, agents often stop exploring before they see the full sequence.  
Third, models often fail to revise early, incorrect beliefs even when faced with subsequent contradictory evidence. 
These results show that simply acquiring visual evidence is not enough. 
Agents must also be able to build and update a reliable perceptual state.
}

\begin{document}

\maketitle

\teaser{\includegraphics[width=0.8\textwidth]{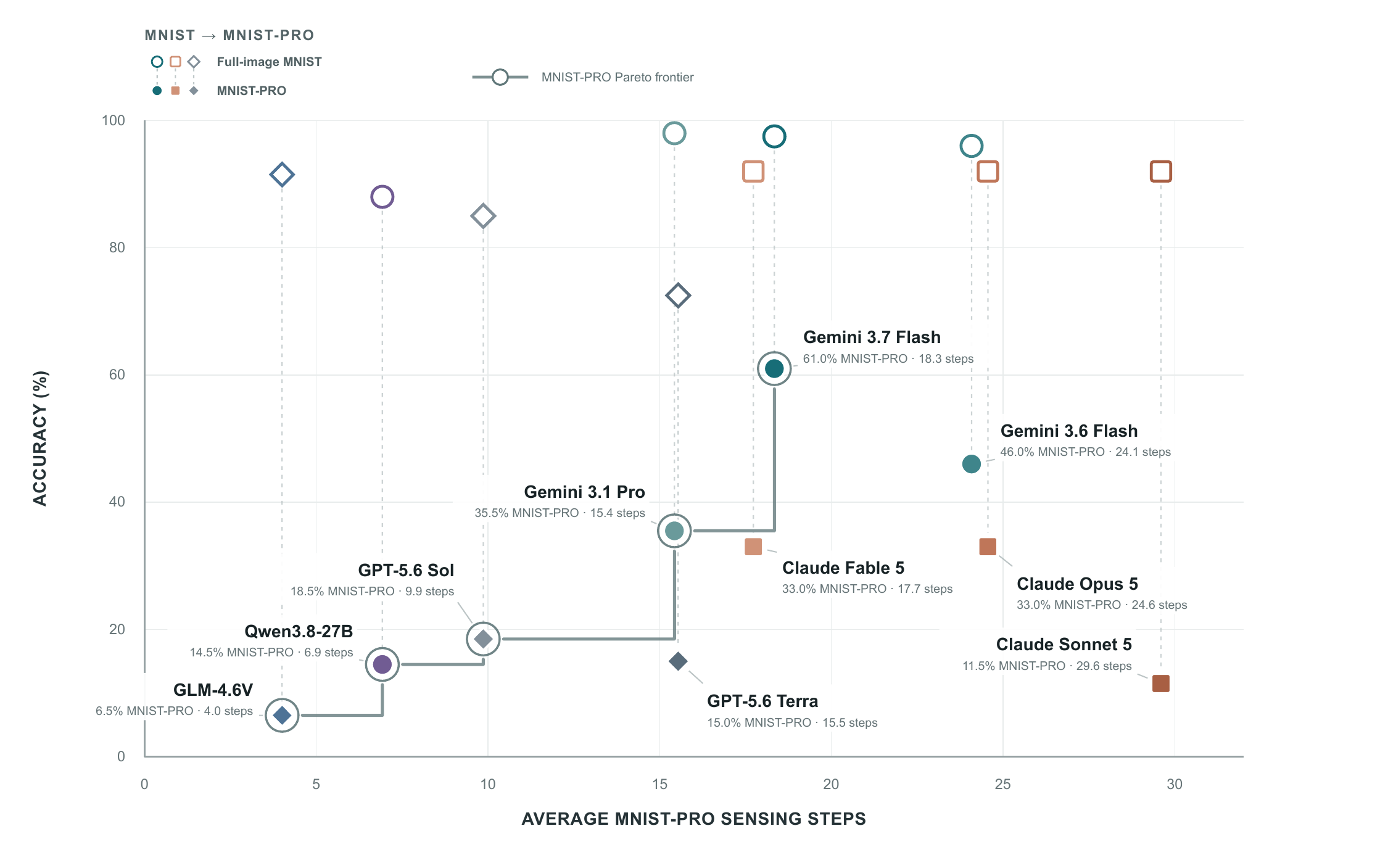}}{}

\section{Introduction}
In this work, we study agentic perception, an increasingly relevant capability as AI agents navigate complex tasks like graphical user interface (GUI) interaction and robotic manipulation \citep{brohan2023rt2visionlanguageactionmodelstransfer,zhou2024webarenarealisticwebenvironment,xie2024osworldbenchmarkingmultimodalagents}.
In agentic perception, AI agents actively gather information from their observations to construct an evolving perceptual state for decision-making.
This differs from the classical active vision literature that has traditionally emphasized controlling sensing to acquire informative observations \citep{bajcsy2016revisitingactiveperception}.
As such, active vision can be viewed as one component of agentic perception.
In partially observable environments, agents must coordinate active visual sensing with working memory to construct a persistent perceptual state of the world.
At any given instant, the agent's visual perception is limited by its field of view or sensor position \citep{savva2019habitatplatformembodiedai}.
To make optimal decisions under this uncertainty, the agent must continuously update its internal representation: it must decide where to look next and consolidate new glimpses with prior observations \citep{wu2023vguidedvisualsearch,xu2026spatialbenchbenchmarkingmultimodallarge,yang2025thinkingspacemultimodallarge}.

Evaluating how effectively multimodal agents construct these perceptual states is challenging due to the confounding factors in existing benchmarks.
First, standard multimodal reasoning benchmarks, such as MMBench \citep{liu2024mmbenchmultimodalmodelallaround} and MMMU \citep{yue2024mmmumassivemultidisciplinemultimodal}, provide the relevant visual inputs together with each question upfront, and therefore do not require models to actively gather or select visual evidence during inference.
Second, active control benchmarks, such as ActiView \citep{wang2025activiewevaluatingactiveperception} and ActiveVision \citep{zhang2026examactiveobservers}, require models to iteratively acquire and reason over multiple visual observations. 
However, these settings do not necessarily distinguish between agents that maintain a coherent, task-relevant perceptual state and agents that rely heavily on revisiting or retaining low-level visual observations. 
This makes it difficult to isolate whether performance reflects genuine stateful perceptual synthesis or effective use of the available visual history.
Finally, while embodied 3D benchmarks such as ALFRED \citep{shridhar2020alfredbenchmarkinterpretinggrounded} require sequential decision-making, their reliance on simulated navigation and object interaction introduces low-level execution failures, such as collisions and unsuccessful interaction actions, that can confound the attribution of failures to perception, memory, planning, or control.

To isolate and evaluate the ability to construct perceptual states under partial observability, we introduce \method{}, a controlled benchmark for sequential perception based on the classical MNIST dataset \citep{726791}.
We formulate active perception as a Partially Observable Markov Decision Process (POMDP) where the agent receives localized visual glimpses of the original image instead of the full image.
Because state-of-the-art vision-language models already achieve near-perfect digit recognition under full observability on MNIST, our setup establishes a high performance ceiling for basic visual perception.
This allows us to largely control for low-level recognition ability and inject difficulty through partial observability.
Unlike complex 3D environments, our 2D canvas removes physical distractions, making it easier to measure the coverage of the original image and track exploration paths.

A key advantage of \method{} is that it enables us to systematically control visual lookback and compare different memory representations for carrying perceptual information forward.
We evaluate models under diverse memory representations, including raw visual history $\hist$, free-form textual thoughts, structured metric grid maps, and a consolidated visual canvas.
By limiting visual lookback, we can additionally expose the memory write-path, forcing agents to serialize fleeting visual evidence into a persistent state before old frames are discarded.
Furthermore, we scale the spatial memory load and search horizon by structuring the benchmark into a Single-Digit task (Level 1) and a Multi-Digit Sequence task (Level 2).
Unlike benchmarks such as ARC-AGI-3 \citep{foundation2026arcagi3newchallengefrontier}, which introduce novel environments with unspecified rules and goals and require agents to infer their dynamics through exploration, \method{} operates in a familiar visual domain with known action semantics, ensuring that failures highlight limitations in agentic perception rather than task comprehension.

In summary, we ask the following research questions:

\begin{enumerate}
    \item \textbf{RQ1:} Can current multimodal agents construct effective perceptual states from sequential, partial observations?

    \item \textbf{RQ2:} How does the representation of perceptual state affect an agent's ability to perceive and act under partial observability?

    \item \textbf{RQ3:} How does perceptual-state construction scale with increasing spatial and temporal demands?

    \item \textbf{RQ4:} Can agents allocate sensing effort productively and determine when the resulting perceptual state is ready for decision-making?
\end{enumerate}

Our experiments reveal four main findings. First, strong visual recognition performance on MNIST does not translate into effective agentic perception on \method{}. Current multimodal agents often struggle to construct a coherent perceptual state by integrating the fragmented visual evidence collected during exploration of the \method{} world. Even when they successfully construct a perceptual state, interpreting that state and making the correct decision from it can remain a bottleneck.

Second, the representation of the perceptual state plays a critical role. We find that retaining raw visual history from explored regions, maintaining textual states, constructing metric grid maps, and programmatically consolidating observations into a visual canvas lead to markedly different results, with no single representation consistently dominating across multimodal agents. Our visual consolidation experiments suggest that the timing of state construction may itself matter: continuously consolidating evidence after each exploration step can be less effective than first exploring without visual consolidation and constructing the consolidated representation only at the end. We hypothesize that repeatedly exposing an incomplete perceptual-state representation may confuse the agent, whereas consolidation becomes useful once sufficient evidence has been acquired.

Third, as we increase the spatial and temporal demands of \method{} by moving from a single digit to a partially observable sequence of two digits whose order must also be preserved, agents struggle significantly more with perceptual-state construction. Moreover, some agents gather substantial visual evidence from both digits but still fail to translate this evidence into correct final predictions. This indicates that increasing task horizon introduces difficulties not only in evidence acquisition, but also in maintaining and interpreting the constructed perceptual state.

Fourth, agents frequently commit to predictions well before exhausting their available sensing budget, suggesting that failures in agentic perception can arise not only from imperfect state construction but also from deciding under unresolved perceptual uncertainty. At the same time, using more of the sensing budget alone does not guarantee success, indicating that agents must determine both \emph{where} to gather additional evidence and \emph{when} the resulting perceptual state is sufficiently reliable for decision-making.

Together, these results indicate that agentic perception depends not only on acquiring informative observations, but also on how those observations are represented and integrated, how effectively the resulting perceptual state can be interpreted, and when the agent decides that sufficient evidence has been acquired to act.

Overall, our contributions are as follows:

\begin{enumerate}
    \item We introduce \method{}, a minimal, controlled benchmark for agentic perception that transforms MNIST from a fully observed digit recognition task into a partially observable world in which agents must explore the environment to actively acquire, integrate, retain, and interpret visual observations over time for decision-making. 
    \item While current multimodal agent evaluations primarily measure task success, \method{} provides a controlled setting to diagnose bottlenecks in agentic perception under partial observability: whether agents fail to acquire sufficient evidence, fail to consolidate the acquired evidence into an effective perceptual-state representation, or fail to correctly interpret the constructed perceptual state.
    \item We develop a controlled evaluation framework for comparing representations of the constructed perceptual states such as raw visual history, free-form textual states, structured metric grid maps, and consolidated visual canvases while systematically varying visual lookback, sensing capacity, and task horizon.
    \item Our evaluation reveals a substantial gap between visual recognition in a fully observable world and agentic perception in a partially observable world. Agents that reliably recognize fully observable digits struggle to consolidate evidence from sequential observations. Additionally, their performance depends largely on how the visual evidence is represented.
    \item We characterize how multimodal agents allocate their sensing or exploration budget and make stopping decisions under partial observability, revealing that agents often commit after limited exploration despite having substantial sensing capacity remaining.
\end{enumerate}

\section{\methodname{}}
\label{sec:activeglimpse_framework}

\begin{figure}
    \centering
    \includegraphics[width=0.99\linewidth]{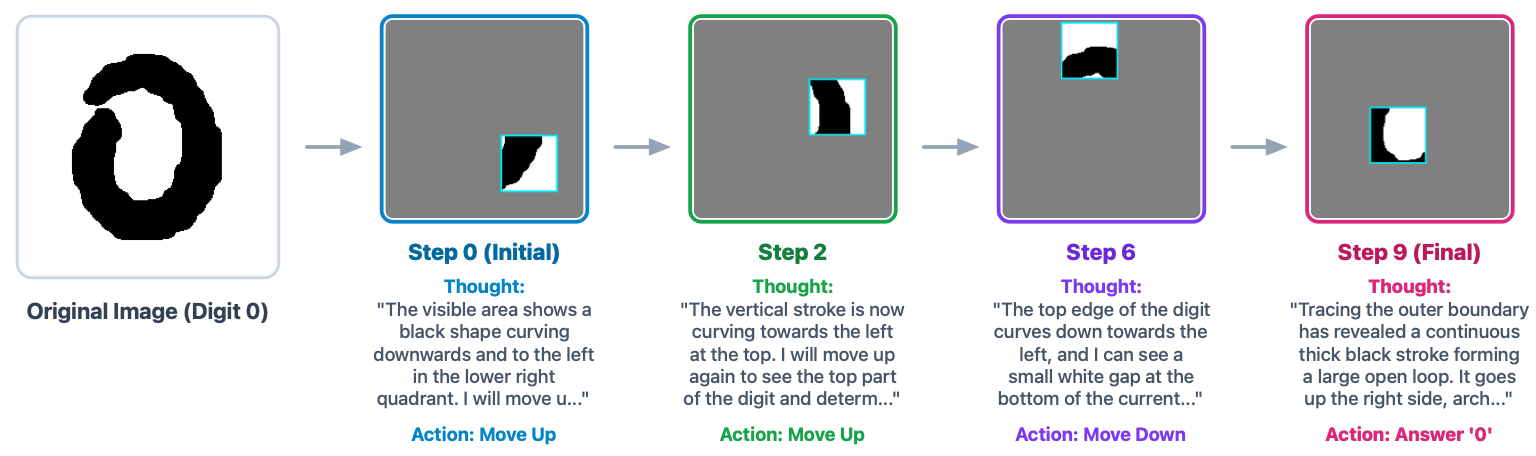}
    \caption{\textbf{Visual exploration and state tracking in \method{}.} The leftmost panel shows the original image (Digit 0), which remains hidden from the agent and can only be accessed through a glimpse window. To solve the task, the agent sequentially moves the glimpse window across the image, updating its beliefs as it gathers new partial observations. At Step 0, the agent begins with a randomly sampled glimpse, stores the observation in memory, and selects an action. It moves upward at Steps 0 and 2, then downward at Step 6. By Step 9, after tracing the digit's boundaries and integrating the accumulated observations, the agent predicts the label (\emph{Answer `0'}).}
    \label{fig:method}
\end{figure}

\method{} evaluates how multimodal agents coordinate sequential, glimpse-based evidence with working memory under partial observability.
At each step, the agent must decide where to look and what information to retain in memory to inform subsequent actions (\Cref{fig:method}).
In this section, we formalize agentic perception as a POMDP (\Cref{sec:pomdp}), define the required agentic loop to isolate sources of agent failure (\Cref{sec:agentic_loop}), introduce a hierarchical task structure (\Cref{sec:levels}), and outline a taxonomy of memory configurations for evaluation (\Cref{sec:taxonomy}).

We instantiate the environment in \method{} on MNIST images with a fixed-size glimpse window that masks the image, revealing only a small region at each step.
This environment isolates spatial tracking and memory without additional visual noise or rendering complexity. 
Performance therefore mainly reflects how well agents explore the environment, combine successive glimpses, and remember what they have seen to solve the task.

\subsection{POMDP Formulation}
\label{sec:pomdp}

We formulate agentic perception as a Partially Observable Markov Decision Process (POMDP) to model sequential visual observations and belief-state tracking under strict sensing budgets.

\paragraph{State Space ($\mathcal{S}$):} 
The state space is the Cartesian product $\mathcal{S} = \mathcal{I} \times \mathcal{P}$, where $\mathcal{I}$ is the set of global canvases $I$ of height $H$ and width $W'$, and $\mathcal{P} = \{0, 1, \dots, W' - w\} \times \{0, 1, \dots, H - h\}$ is the coordinate space for the top-left corner of the agent's glimpse window. 
At each timestep $t$, the underlying state $s_t = (I, p_t) \in \mathcal{S}$ consists of the static canvas $I$ and the agent's coordinates $p_t = (x_t, y_t) \in \mathcal{P}$.

\paragraph{Observation Space ($\mathcal{O}$):} 
At each timestep $t$, the agent receives a masked observation $o_t \in \mathcal{O}$ of size $H \times W'$. 
Only a glimpse window of size $h \times w$ (where $h < H$ and $w < W'$) covering the region $[x_t, x_t + w) \times [y_t, y_t + h)$ is visible, while all regions outside this window are masked. 
To observe the full canvas, the agent must explore the environment sequentially.

\paragraph{Action Space ($\mathcal{A}$):} 
The action space is divided into movement and prediction actions, such that $\mathcal{A} = \mathcal{A}_{\text{move}} \cup \mathcal{A}_{\text{predict}}$. 
Movement actions $a \in \mathcal{A}_{\text{move}} = \{\text{up}, \text{down}, \text{left}, \text{right}\}$ shift the coordinates $p_t$ in the cardinal directions by a step size $\delta$. 
Prediction actions $a \in \mathcal{A}_{\text{predict}}$ terminate the episode and produce a prediction from the label space $\mathcal{Y}$, where $\mathcal{Y} = \{0, \dots, 9\}$ for Level 1 and $\mathcal{Y} = \{0, \dots, 9\}^N$ for Level 2 (representing a sequence of $N$ concatenated digits).

\paragraph{Transition Function ($\mathcal{T}$):} 
The transition function $\mathcal{T}: \mathcal{S} \times \mathcal{A} \rightarrow \mathcal{S}$ governs the dynamics of the environment. 
A movement action $a_t \in \mathcal{A}_{\text{move}}$ updates the coordinates to $p_{t+1} = \text{clip}(p_t + \delta \cdot u(a_t))$, where $\delta$ is step size and $u(a_t)$ is the unit displacement vector: $u(\text{up}) = (0, -1)$, $u(\text{down}) = (0, 1)$, $u(\text{left}) = (-1, 0)$, and $u(\text{right}) = (1, 0)$. 
The clipping function bounds $p_{t+1}$ within the coordinate space $\mathcal{P}$. A prediction action $a_t \in \mathcal{A}_{\text{predict}}$ transitions the environment to a terminal state.

\paragraph{Observation Function ($\Omega$):} 
The observation function $\Omega: \mathcal{S} \rightarrow \mathcal{O}$ maps the underlying state $s_t = (I, p_t)$ to the masked observation $o_t$ that reveals only the pixels in window $[x_t, x_t + w) \times [y_t, y_t + h)$ of the global canvas $I$, while the remaining pixels stay masked with a constant gray value.

Since the underlying state $s_t$ is only partially observable, the agent cannot make informed decisions based solely on the current observation $o_t$. 
Instead, it must retain and update information from past observations and actions to construct a perceptual state, which serves as a running estimate of the environment's underlying state.

The environment terminates the episode when the agent makes a prediction or the step budget $T_{\max}$ is reached, preventing infinite loops. 
To ensure this budget is sufficient, we define $T_{\max}$ as the number of discrete steps needed to cover the canvas with a stride length of $\delta$:
\begin{equation}
    \label{eq:tmax}
    T_{\max} = \text{steps}_{\text{width}} \times \text{steps}_{\text{height}}
\end{equation}
where:
\begin{align}
    \text{steps}_{\text{width}} &= \left\lceil \frac{\max(0, W' - w)}{\delta} \right\rceil + 1 \\
    \text{steps}_{\text{height}} &= \left\lceil \frac{\max(0, H - h)}{\delta} \right\rceil + 1
\end{align}
and $W'$ is the active canvas width. In Level 1, the canvas is a square ($W' = W$ and $H = W$), whereas in Level 2, the canvas contains $N$ horizontally concatenated digits ($W' = N \cdot W$).

\subsection{The Required Agentic Loop}
\label{sec:agentic_loop}
\begin{figure}
  \centering
  \vspace{-0.5\baselineskip}
  \includegraphics[width=\linewidth]{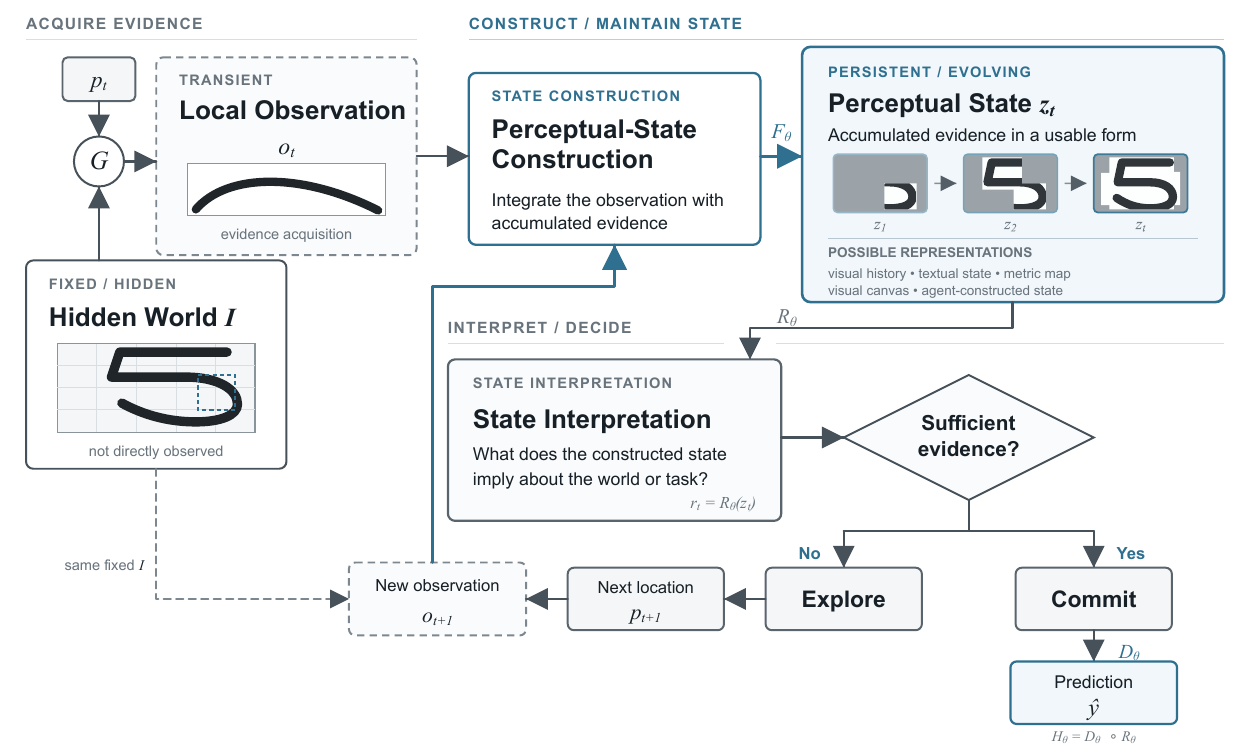}
  \caption{Perceptual-state construction under partial observability. Evidence acquisition, state construction, and state interpretation are distinct parts of the workflow. The agent explores until it commits to a prediction.}
  \label{fig:perceptual_state_loop}
  \vspace{-0.5\baselineskip}
\end{figure}

Solving the task under the \method{} setup requires a straightforward agentic loop. 
This simple design is deliberate, as it helps identify the required abilities to solve the problem and, consequently, locate missing abilities. 
We present the agentic loop in \Cref{fig:perceptual_state_loop}. 
Let $I$ denote the fixed hidden world and $p_t$ be the agent's sensing location at timestep $t$. 
The agent observes $o_t = \Omega(I,p_t)$, and updates its perceptual state as $z_t = F_\theta(z_{t-1}, o_t, p_t)$. To distinguish perceptual-state construction from its subsequent interpretation, we further factor the prediction function $H_\theta$ as $H_\theta = D_\theta \circ R_\theta$, where $r_t = R_\theta(z_t)$ represents the interpretation or readout of the constructed perceptual state and $\hat{y} = D_\theta(r_t)$ produces the task prediction. Based on $z_t$, the agent chooses an action $a_t \sim \pi_\theta(\cdot \mid z_t)$. The agent decides whether $a_t$ is an exploration or a prediction $\hat{y} = H_\theta(z_t)$. If $a_t$ is an exploration, the sensing location is updated to $p_{t+1} = T(p_t,a_t)$ and the observation is updated to $o_{t+1}$.

\subsection{Benchmark Task Levels and Complexity}
\label{sec:levels}

We evaluate agents across two hierarchical task levels of increasing difficulty. 
The two levels are designed to progressively test spatial tracking, visual integration, long-horizon search, and sequential memory. 
Level 1 focuses on tracking information within a single coordinate frame, while Level 2 extends this setting to multiple concatenated frames that must be processed and retained in the correct temporal order.

\begin{enumerate}
    \item \textbf{Level 1: Spatial Integration (Single Digit):} 
    The agent is presented with a MNIST digit and must trace and identify the single centered digit. 
    This level provides a baseline measure of the agent's ability to track spatial coordinates and integrate spatially disjoint visual features into a coherent representation of the digit.

    \item \textbf{Level 2: Sequential Ordering and Tracking (Multi-Digit):} 
    We concatenate $N$ independent digits horizontally on a canvas. The agent must search across the canvas, locate and recognize each digit, and preserve their relative order throughout the sequence (e.g., mapping the visual sequence to \texttt{58}). This level introduces the additional challenges of long-horizon spatial search, sequential retention, and order tracking across multiple visual frames.
\end{enumerate}

The difficulty of both levels can be controlled by varying the observation window dimensions $h \times w$ and the step size $\delta$. 
Reducing the window size increases the trajectory length required to cover the full canvas, increasing the amount of spatial information that must be retained over time. 
This provides a way to increase the memory demands of the benchmark while keeping the underlying dataset unchanged.

\subsection{A Taxonomy of Memory for Multimodal Agents}
\label{sec:taxonomy}

To analyze how different memory architectures support perceptual state construction under partial observability, we evaluate agents across five core memory configurations classified into three primary categories: internal state construction, externalized state construction, and procedural memory.
This structured classification allows us to isolate the specific performance bottlenecks associated with textual, spatial, and visual working memory.
While our main benchmark evaluation focuses on three internal configurations (\textit{Image Only Baseline}, \textit{Textual State}, and \textit{Metric Grid Map}), we also introduce the \textit{Visual Memory Canvas} (evaluated in both online and offline modes) as an analytical tool to ablate the limits of combining spatial information derived from partial visual inputs.
Finally, we examine an agentic tool-use harness with persistent episodic memory to study procedural skill retention.

Overall, the perceptual state representation and utilization of memory to solve \method{} are categorized into three primary categories:

\subsubsection{Internal State Construction}
\begin{enumerate}
    \item \textbf{Image Only Baseline:} 
    This configuration provides the agent with a chronological text log of its past actions alongside the full visual history $\hist$. 
    Previous actions are recorded in text form:
\begin{lstlisting}[language=json]
{
    "action": "move", 
    "direction": "right"
}
\end{lstlisting}
    The agent does not maintain a written state or summary of its visual observations.
    Instead, it receives the chronological sequence of past visual glimpses and actions.
    This baseline isolates the model's ability to perform path integration and spatial reasoning directly from visual history, without relying on any intermediate textual representation.

    \item \textbf{Textual State:} 
    This configuration requires the agent to generate a \texttt{"thought"} alongside its chosen action. 
    These thoughts are accumulated in the working memory, forming a running textual state. 
    Under a strict zero-visual-lookback constraint ($\hist=1$), the agent has no access to past frames and must rely entirely on this textual state to coordinate its future movements.
    This configuration evaluates the model's ability to maintain a free-form, qualitative representation of the environment over time.
    At each turn, the agent outputs its reasoning and action in a JSON format, as shown below:
\begin{lstlisting}[language=json]
{
    "thought": "I see a curved stroke in the center. I will move down to trace the loop.",
    "action": "move",
    "direction": "down"
}
\end{lstlisting}

    \item \textbf{Metric Grid Map:} 
    This configuration requires the agent to generate a structured spatial state on top of its \texttt{"thought"} and chosen action, representing its current location on a 2D grid.
    These spatial states are stored in memory, forming a structured coordinate map that the agent can use to reconstruct the environment over time.
    Under a strict zero-visual-lookback constraint ($\hist=1$), the agent has no access to past frames and must rely on this structured spatial memory to coordinate its future decisions.
    This evaluates the model's ability to maintain a geometrically grounded representation through coordinate anchoring, path integration, and feature mapping.

    The spatial memory is formalized as follows:
    
    \textbf{Coordinate Anchoring:} At the initial timestep $t=0$, the agent anchors its starting position as the origin of its relative coordinate frame:
    \begin{equation}
        \mathbf{x}_0 = [0, 0].
    \end{equation}

    \textbf{In-Context Path Integration:} At each step $t > 0$, the agent updates its current relative coordinate $\mathbf{x}_t$ by adding the displacement associated with its previous movement action $a_{t-1}$ to the previous coordinate:
    \begin{equation}
        \mathbf{x}_t = \mathbf{x}_{t-1} + \Delta(a_{t-1}),
    \end{equation}
    where the displacement vectors are defined as
    $\Delta(\text{\texttt{up}}) = [0,-1]$,
    $\Delta(\text{\texttt{down}}) = [0,1]$,
    $\Delta(\text{\texttt{left}}) = [-1,0]$, and
    $\Delta(\text{\texttt{right}}) = [1,0]$.

    \textbf{Structured JSON Output:} At each turn, the model is constrained to output a structured JSON object containing its reasoning (\texttt{"thought"}), its next action, and its current observation mapped to the computed coordinate under the key \texttt{"spatial\_map"}:
\begin{lstlisting}[language=json]
{
  "thought": "I see empty space at coordinate [0, 0]. I will move right.",
  "spatial_map": [
    {
      "coords": [0, 0],
      "features": "empty background"
    }
  ],
  "action": "move",
  "direction": "right"
}
\end{lstlisting}

    \textbf{Chronological Memory Consolidation:} 
    Because the conversation history preserves the model's past structured outputs, the accumulated spatial map $\mathcal{M}$ is implicitly maintained across turns within the model's context window. 
    At each subsequent turn $T$, the agent receives the chronological sequence of previous actions, thoughts, and structured spatial maps:
    \begin{equation}
        \mathcal{M} = \big\{ (\mathbf{x}_t, \mathbf{f}_t) \big\}_{t=0}^{T-1},
    \end{equation}
    where $\mathbf{f}_t$ denotes the observed features at coordinate $\mathbf{x}_t$. 
    The agent references this structured history to maintain a consistent spatial representation across turns, without requiring a separate external database.

    \end{enumerate}

    \subsubsection{Externalized State Construction}
    To isolate high-level reasoning and diagnose where memory bottlenecks occur, we explore two distinct modes of visual memory consolidation:
    \begin{enumerate}
    \item \textbf{Online Visual Memory Canvas:} 
    To isolate high-level visual reasoning from the difficulty of reconstructing spatial information across sequential observations, we programmatically consolidate all past glimpses into a single persistent visual canvas. 
    Instead of providing the model with a sequential history of separate glimpses, the environment maintains a single coordinate-aligned image $C_t$ that accumulates all observations over time. 
    At $t=0$, $C_0$ is initialized with a gray background, and the current glimpse is placed at its corresponding location on the canvas. 
    At each subsequent step $t>0$, the newly observed glimpse is overlaid onto the previous canvas $C_{t-1}$ at its exact canvas coordinates $p_t$:
    \begin{equation}
        C_t(u, v) = \begin{cases}
        I(u, v) & \text{if } (u, v) \in [x_t, x_t + w) \times [y_t, y_t + h) \\
        C_{t-1}(u, v) & \text{otherwise}.
        \end{cases}
    \end{equation}
    This setup preserves the spatial relationships between observations and eliminates the need for the model to align and integrate multiple historical images.
    To distinguish the accumulated visual context from the agent's most recent observation, the model receives two images at each step $t$: (1) the consolidated canvas $C_t$, which captures the accumulated spatial context, and (2) the latest glimpse $G_t$, which provides the agent's current location. 
    This dual-image representation keeps the amount of visual input constant at $\mathcal{O}(1)$, while the environment handles the spatial alignment and fusion of historical observations.

    \item \textbf{Offline Visual Memory Canvas:} Similarly, we construct an offline visual memory canvas from the historical sequence of glimpses collected during Image Only and Textual State experiments. 
    This serves as an analytical probing stage to isolate representation quality from sequential planning.
    The consolidated canvas is presented to the agent to make a final prediction, allowing us to diagnose if failures stem from exploration or interpretation.
\end{enumerate}

\subsubsection{Persistent Memory}
\begin{enumerate}
    \item \textbf{Harness Systems.} We further evaluate whether \method{} remains challenging in a realistic agentic setting where agents have the autonomy to use tools while solving the task. In particular, we examine whether agents autonomously discover useful strategies, such as assembling sequential glimpses into a visual canvas. Such strategies are expected to be explored purely autonomously and not just limited to canvas building; e.g., we notice agents autonomously decide whether to feed one or multiple observations to predict. We provide persistent memory across episodes to test whether agents can retain successful procedures or skills and reuse them in future episodes. If they contain reusable procedures, we call them \emph{procedural memory}. In addition, feedback on the correctness of the agent's decisions is stored in memory, allowing us to study whether agents revise previously learned strategies or beliefs in response to outcome feedback.
    
\end{enumerate}

\section{Experiments and Results}

We first compare full-image recognition with performance from sequential glimpses. We then examine the trajectories and replay them with alternative state representations to see where errors arise. The harness experiments test whether models discover and reuse their own procedures. Finally, we vary visual history and glimpse size.

\subsection{Experimental Setup}

\paragraph{Dataset Preprocessing.}
For each $28 \times 28$ MNIST digit image, we invert the grayscale values so that the digit strokes appear black on a white background.
We then upscale the images to $224 \times 224$ pixels using bilinear interpolation.
To remove the blur from upscaling, we binarize the images with an intensity threshold of 200, so that each pixel is either 0 (black stroke) or 255 (white background).
These binary images are the inputs for the Level 1 single-digit task.
For the Level 2 multi-digit task, we horizontally concatenate two raw digit images before applying this same pipeline, which yields a combined image of $224 \times 448$ pixels.

\paragraph{Models.}
We evaluate ten models, including both proprietary and open-source.
The proprietary models are \geminipro{}, \geminisix{}, \geminiseven{}, \claudesonnet{}, \claudeopus{}, \claudefable{}, \gptterra{}, and \gptsol{}.
The open-source models are \qwen{} \citep{qwen3.8} and \glm{} \citep{vteam2025glm45vglm41vthinkingversatilemultimodal}.
We query all models using their default API configurations, including default temperatures and thinking levels.
To study the impact of scaling reasoning budgets, we also evaluate \gptsol{} under \textit{xhigh} reasoning effort configuration (denoted as \gptsolxhigh{}) for selected experiments.

\paragraph{Episode and Environment Settings.}
We evaluate the models on 100 test episodes for both tasks under a partially observable active vision setup.
For Level 1, the episodes are balanced across the ten digit classes, with ten episodes per class.
For Level 2, each episode has two digits with labels sampled independently and uniformly from $\{0, 1, \dots, 9\}$.
At the start of an episode, the environment masks the entire image except for a random glimpse window of $64 \times 64$ pixels.
The agent navigates this window using movement actions $\mathcal{A}_{\text{move}}$ with a step size of $\delta = 32$ pixels.
Each episode ends when the agent makes a prediction $\mathcal{A}_{\text{predict}}$ or when it exceeds its step budget.
Based on \Cref{eq:tmax}, the step budget is 36 steps for Level 1 and 78 steps for Level 2.

\paragraph{Evaluation Metrics.}
We use three metrics to evaluate each model configuration.
First, \emph{control accuracy} is the classification accuracy on the fully unmasked, fully observable image, which provides an upper bound.
Second, \emph{classification accuracy} is the agent's prediction accuracy under partial observability.
Third, the \emph{average step count} is the average number of steps the agent takes before predicting, which indicates its exploration efficiency.

\begin{table*}[t]
\centering
%\small
\resizebox{0.9\linewidth}{!}{%
\setlength{\tabcolsep}{5.5pt}
\renewcommand{\arraystretch}{1.12}
\begin{tabular}{lccccccccc}
\toprule
\multirow{2}{*}{\textbf{Model}}
& \multicolumn{3}{c}{\textbf{Control Accuracy (\%) $\uparrow$}}
& \multicolumn{3}{c}{\textbf{Multi-turn Accuracy (\%) $\uparrow$}}
& \multicolumn{3}{c}{\textbf{Average Steps}} \\
\cmidrule(lr){2-4}
\cmidrule(lr){5-7}
\cmidrule(lr){8-10}
& L1 & L2 & \textbf{Avg.}
& L1 & L2 & \textbf{Avg.}
& L1 & L2 & \textbf{Avg.} \\
\midrule
\multicolumn{10}{l}{\textit{Proprietary Models}} \\
\geminipro{}
& 99.0 & 97.0 & 98.0
& 53.0 & 18.0 & 35.5
& 7.33 & 23.53 & 15.43 \\
\geminisix{}
& 96.0 & 96.0 & 96.0
& 64.0 & 28.0 & 46.0
& 9.48 & 38.69 & 24.09 \\
\geminiseven{}
& 98.0 & 97.0 & 97.5
& 75.0 & 47.0 & 61.0
& 8.23 & 28.45 & 18.34 \\
\claudesonnet{}
& 94.0 & 90.0 & 92.0
& 23.0 & 0.0 & 11.5
& 15.96 & 43.24 & 29.60 \\
\claudeopus{}
& 93.0 & 91.0 & 92.0
& 49.0 & 17.0 & 33.0
& 12.26 & 36.86 & 24.56 \\
\claudefable{}
& 95.0 & 89.0 & 92.0
& 50.0 & 16.0 & 33.0
& 10.45 & 25.01 & 17.73 \\
\gptterra{}
& 83.0 & 62.0 & 72.5
& 30.0 & 0.0 & 15.0
& 10.29 & 20.79 & 15.54 \\
\gptsol{}
& 88.0 & 82.0 & 85.0
& 37.0 & 0.0 & 18.5
& 5.74 & 13.99 & 9.87 \\
\addlinespace[2pt]
\multicolumn{10}{l}{\textit{Open-Source Models}} \\
\qwen{}
& 89.0 & 87.0 & 88.0
& 29.0 & 0.0 & 14.5
& 4.62 & 9.23 & 6.93 \\
\glm{}
& 91.0 & 92.0 & 91.5
& 13.0 & 0.0 & 6.5
& 2.76 & 5.25 & 4.01 \\
\bottomrule
\end{tabular}
}
\caption{Performance on \method{} with native multi-turn conversation history.}
\label{tab:natural_multiturn}
\end{table*}

\subsection{Native Multi-turn Setup}
First, we create a natural multi-turn setup for the agents. This setup does not allow tool use, as we are interested in systematically ablating the use of tools. Agents are first given a user prompt describing the task along with the first glimpse. The agent responds with an action, and in the next turn receives only an image from the user without any additional prompt. The goal is to create a setup in which the agent explores based on the initial user instruction, discovers the world, and continues acting. The full conversation trace, including all explored images, thoughts, and actions, remains available to the agent for future actions in this setup.

We present the results in \Cref{tab:natural_multiturn}.
\geminiseven{} achieves the highest accuracy (75.0\% on Level 1 and 47.0\% on Level 2), followed by \geminisix{} (64.0\% and 28.0\%) and \geminipro{} (53.0\% and 18.0\%).
The Claude models explore more extensively, but they struggle to translate their exploration into correct predictions.
\claudesonnet{} has the highest step counts (15.96 on Level 1 and 43.24 on Level 2), but achieves only 23.0\% accuracy on Level 1 and 0.0\% on Level 2.
Similarly, \claudeopus{}, \claudefable{}, and \gptterra{} use higher average steps but achieve moderate or poor accuracy.
For models such as \gptsol{}, \qwen{}, and \glm{}, they rush to predict in very few steps.
They average 5.74, 4.62, and 2.76 steps on Level 1, and 13.99, 9.23, and 5.25 on Level 2.
This early stopping means they often make a prediction before gathering enough information.
On Level 2, this leads to 0.0\% accuracy, and on Level 1, they achieve 37.0\%, 29.0\%, and 13.0\% accuracy, respectively.

The variables represented in \Cref{fig:perceptual_state_loop} are not independently controlled in this setup, making it difficult to determine whether a failure arises from evidence acquisition, perceptual-state construction, interpretation of the accumulated evidence, or the decision to commit. Analysis of the interaction traces reveals that the failures are primarily caused by misinterpreting the accumulated evidence. We therefore use the targeted experimental conditions described next to isolate these components.

\subsection{Results of Different Internal State Construction Mechanisms}

\begin{table*}[t]
\centering
\resizebox{\linewidth}{!}{%
\begin{tabular}{lccc|cccccc|cccccc|cccccc}
\toprule
\multirow{3}{*}{\textbf{Model}}
& \multicolumn{3}{c}{\textbf{Control Acc.}}
& \multicolumn{6}{c}{\textbf{Image Only} ($\hist{=}\infty$)}
& \multicolumn{6}{c}{\textbf{Textual State} ($\hist{=}1$)}
& \multicolumn{6}{c}{\textbf{Metric Grid Map} ($\hist{=}1$)} \\
\cmidrule(lr){2-4} \cmidrule(lr){5-10} \cmidrule(lr){11-16} \cmidrule(lr){17-22}
& \multicolumn{3}{c}{\textbf{(\%) $\uparrow$}}
& \multicolumn{3}{c}{\textbf{Acc. (\%) $\uparrow$}} & \multicolumn{3}{c}{\textbf{Avg. Steps $\downarrow$}}
& \multicolumn{3}{c}{\textbf{Acc. (\%) $\uparrow$}} & \multicolumn{3}{c}{\textbf{Avg. Steps $\downarrow$}}
& \multicolumn{3}{c}{\textbf{Acc. (\%) $\uparrow$}} & \multicolumn{3}{c}{\textbf{Avg. Steps $\downarrow$}} \\
\cmidrule(lr){2-4} \cmidrule(lr){5-7} \cmidrule(lr){8-10} \cmidrule(lr){11-13} \cmidrule(lr){14-16} \cmidrule(lr){17-19} \cmidrule(lr){20-22}
& L1 & L2 & \textbf{Avg} & L1 & L2 & \textbf{Avg} & L1 & L2 & \textbf{Avg}
& L1 & L2 & \textbf{Avg} & L1 & L2 & \textbf{Avg} & L1 & L2 & \textbf{Avg} & L1 & L2 & \textbf{Avg} \\
\midrule
\multicolumn{22}{l}{\textit{Proprietary Models}} \\
\geminipro{}    & 99.0 & 97.0 & 98.0 & 38.0 & \phantom{0}5.0 & 21.5 & 4.52 & 11.95 & \phantom{0}8.23 & 56.0 & 20.0 & \textbf{38.0} & 7.18 & 18.28 & 12.73 & 53.0 & 14.0 & 33.5 & 7.15 & 19.19 & 13.17 \\
\geminisix{}    & 96.0 & 96.0 & 96.0 & 53.0 & 18.0 & \textbf{35.5} & 5.62 & 15.11 & 10.37 & 47.0 & 15.0 & 31.0 & 9.35 & 39.29 & 24.32 & 49.0 & 10.0 & 29.5 & 7.92 & 20.44 & 14.18 \\
\geminiseven{}  & 98.0 & 97.0 & 97.5 & 65.0 & 22.0 & \textbf{43.5} & 5.26 & 14.69 & \phantom{0}9.97 & 54.0 & \phantom{0}6.0 & 30.0 & 7.92 & 34.84 & 21.38 & 57.0 & 28.0 & 42.5 & 6.82 & 20.37 & 13.60 \\
\claudesonnet{} & 94.0 & 90.0 & 92.0 & 24.0 & \phantom{0}1.0 & \textbf{12.5} & 17.35 & 45.15 & 31.25 & 22.0 & \phantom{0}2.0 & 12.0 & 14.79 & 75.64 & 45.22 & 19.0 & \phantom{0}2.0 & 10.5 & 13.07 & 50.39 & 31.73 \\
\claudeopus{}   & 93.0 & 91.0 & 92.0 & 43.0 & 13.0 & \textbf{28.0} & 7.43 & 22.50 & 14.96 & 41.0 & 13.0 & 27.0 & 11.61 & 39.90 & 25.75 & 36.0 & 15.0 & 25.5 & 10.57 & 29.55 & 20.06 \\
\claudefable{}  & 95.0 & 89.0 & 92.0 & 47.0 & 15.0 & 31.0 & 7.27 & 22.30 & 14.79 & 38.0 & 18.0 & 28.0 & 11.30 & 35.93 & 23.62 & 45.0 & 23.0 & \textbf{34.0} & 9.92 & 28.62 & 19.27 \\
\gptterra{}     & 83.0 & 62.0 & 72.5 & 35.0 & \phantom{0}1.0 & \textbf{18.0} & 18.19 & 39.67 & 28.93 & 25.0 & \phantom{0}0.0 & 12.5 & 24.09 & 69.67 & 46.88 & 24.0 & \phantom{0}0.0 & 12.0 & 21.92 & 62.43 & 42.17 \\
\gptsol{}       & 88.0 & 82.0 & 85.0 & 37.0 & \phantom{0}1.0 & \textbf{19.0} & 6.53 & 16.98 & 11.76 & 29.0 & \phantom{0}0.0 & 14.5 & 7.01 & 22.52 & 14.77 & 32.0 & \phantom{0}1.0 & 16.5 & 7.89 & 21.26 & 14.58 \\
\gptsolxhigh{}& 92.0 & 79.0 & 85.5 & 56.0 & 29.0 & \textbf{42.5} & 9.59 & 35.43 & 22.51 & 37.0 & \phantom{0}3.0 & 20.0 & 9.19 & 28.00 & 18.59 & 39.0 & \phantom{0}5.0 & 22.0 & 9.79 & 36.09 & 22.94 \\
\addlinespace[2pt]
\multicolumn{22}{l}{\textit{Open-Source Models}} \\
\qwen{}         & 89.0 & 87.0 & 88.0 & 28.0 & \phantom{0}2.0 & 15.0 & 4.76 & \phantom{0}9.19 & \phantom{0}6.97 & 31.0 & \phantom{0}1.0 & 16.0 & 6.97 & 16.94 & 11.96 & 36.0 & \phantom{0}0.0 & \textbf{18.0} & 6.86 & 13.87 & 10.37 \\
\glm{}          & 91.0 & 92.0 & 91.5 & 14.0 & \phantom{0}0.0 & \phantom{0}7.0 & 3.49 & \phantom{0}6.51 & \phantom{0}5.00 & 20.0 & \phantom{0}1.0 & \textbf{10.5} & 20.24 & 40.08 & 30.16 & 20.0 & \phantom{0}0.0 & 10.0 & 11.39 & 27.67 & 19.53 \\
\bottomrule
\end{tabular}%
}
\caption{Model performance on \method{}. Each configuration reports accuracy and average steps for Level 1 (Single-Digit), Level 2 (Multi-Digit), and their mean. \textbf{Bold} indicates the partially observable configuration with the highest average accuracy for each model.}
\label{tab:main_results}
\end{table*}

\paragraph{The Gap Between Recognition and Perceptual State Construction.}
High recognition accuracy under full observability does not ensure successful perceptual state construction. 
Under the control condition, almost all models identify unmasked digits with 83.0\% to 99.0\% accuracy (\Cref{tab:main_results}).
However, performance drops significantly under partial observability, even when models can look back at all previous glimpses ($\hist=\infty$).
For example, \geminipro{} drops from 99.0\% control accuracy to 38.0\%, and \claudesonnet{} falls from 94.0\% to 24.0\%. 
These performance drops demonstrate that passive classification capability does not necessarily translate into the ability to actively gather and integrate sequential visual observations for downstream decision making.

\paragraph{Memory Representations Help and Hurt Differently.}
Without a visual history ($\hist=1$), models must rely on written memory, using either a \emph{Textual State} or a \emph{Metric Grid Map} to carry information forward.
Adding coordinate tracking affects models differently.
For example, \geminipro{} reaches 56.0\% accuracy with textual reasoning alone, but drops to 53.0\% when required to additionally maintain a grid map.
Conversely, \geminiseven{} benefits from spatial tracking, improving from 54.0\% with textual memory to 57.0\% with the grid map.
Thus, tracking coordinates is not always helpful. 
The added work of path integration can degrade performance for some models, while giving others a stable geometric state construction that improves their predictions.

\paragraph{Textual State Does Not Show a Consistent Pattern.}
For Level 1 (Single-Digit) tasks, keeping a raw visual history of past glimpses ($\hist=\infty$) results in the highest performance, with seven of the eight proprietary models performing best in this image-only setting.
For example, \geminisix{} scores 53.0\% compared to 47.0\% when using textual state, and \geminiseven{} scores 65.0\% compared to 54.0\%.
For Level 2 (Multi-Digit) tasks, however, the direct visual history baseline results in lower performance, for instance, \geminipro{} scores 5.0\% and \claudefable{} scores 15.0\%.
Performance is higher when utilizing textual memory representations, such as textual states or metric grid maps.
With a textual state, \geminipro{} reaches 20.0\%, while the use of metric grid maps results in \claudefable{} and \geminiseven{} reaching 23.0\% and 28.0\%, respectively. However, there is no consistent trend, as different models showed different comparative results between these two state representations.

\paragraph{Challenges of Multi-Digit Sequences.}
Despite the benefits of these memory representations, multi-digit agentic perception remains challenging.
In Level 2, scaling the task to multi-digits leads to a huge drop in performance across almost all models under partial observability. 
In the image-only baseline, most models score below 20.0\% accuracy, for instance, \geminipro{} drops to 5.0\% and \gptsol{} drops to 1.0\%. 
Alternative representations, such as textual states or metric grid maps, yield mixed but generally low results.
For example, \geminipro{} reaches 20.0\% with a textual state, while \geminiseven{} reaches 28.0\% with a metric grid map.
Multi-digit sequences under partial observability remain a major challenge as no model exceeds 30.0\% accuracy, regardless of the memory representation.

\paragraph{Influence of Visual History on Exploration Steps.}
Restricting past visual history often increases the number of exploration steps models take before predicting.
With an unbounded visual history ($\hist=\infty$), models use fewer steps.
For example, \geminipro{} averages 4.52 steps on Level 1.
Without visual lookback ($\hist=1$), trajectories grow longer.
The average step count for \geminipro{} increases to 7.18 steps under a textual state and 7.15 steps under a metric grid map.
These results show that an unbounded visual history generally reduces the average number of steps compared to relying on a textual state or a metric grid map.

\begin{figure*}[t]
  \centering
  \includegraphics[width=0.99\textwidth]{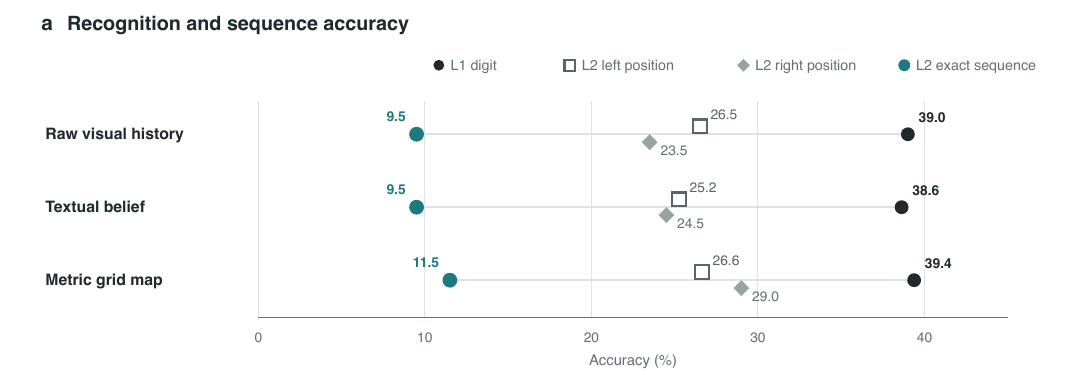}

  \vspace{0.35em}
  \begin{minipage}[t]{0.49\textwidth}
    \vspace{0pt}
    \centering
    \includegraphics[width=\linewidth]{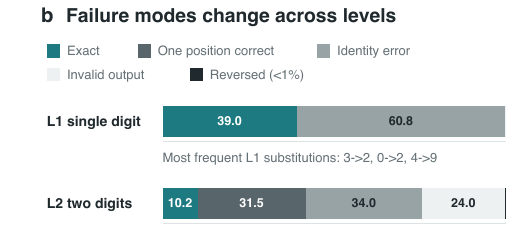}
  \end{minipage}\hfill
  \begin{minipage}[t]{0.49\textwidth}
    \vspace{0pt}
    \centering
    \includegraphics[width=\linewidth]{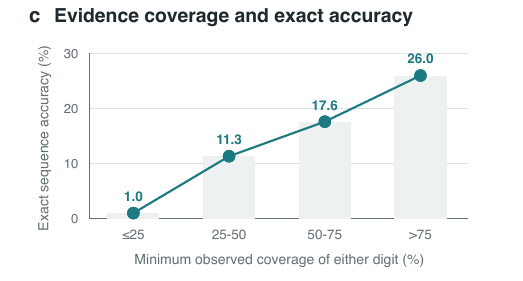}
  \end{minipage}

  \caption{Level 2 introduces search and sequence-construction failures beyond single-digit recognition.}
  \label{fig:mnist-pro-multidigit-failure-analysis}
\end{figure*}

\subsection{Diagnostic Analyses}

\subsubsection{Internal Perceptual-State Failure Modes}

To systematically analyze why agents fail as the task scales, we present a detailed decomposition of recognition accuracy, outcome distributions, and the effect of visual coverage in \Cref{fig:mnist-pro-multidigit-failure-analysis}.

\paragraph{Positional and Exact-Sequence Accuracy (\Cref{fig:mnist-pro-multidigit-failure-analysis}a).}
As shown in \Cref{fig:mnist-pro-multidigit-failure-analysis}a, scaling the task from single digits (Level 1) to multi-digit sequences (Level 2) leads to a performance drop across all three state representations (Raw Visual History, Textual Belief, and Metric Grid Map), despite models maintaining near-perfect control accuracy under full observability. 
While single-digit Level 1 accuracy is roughly 39.0\% across configurations, individual positional accuracy (left and right positions) in Level 2 drops to around 23.5\% to 29.0\%. 
The exact sequence accuracy further drops to around 9.5\% to 11.5\%. 

\paragraph{Failure Mode Shift Across Levels (\Cref{fig:mnist-pro-multidigit-failure-analysis}b).}
The distribution of errors qualitatively shifts as the task scales to multi-digits. 
In Level 1, errors are mainly valid digit substitutions (identity errors) comprising 60.8\% of all trials, with invalid outputs being negligible (0.17\%).
Level 2 shows higher frequency of failures, with 24.0\% of outputs being invalid.
One-position-correct outcomes account for 31.5\%, while exact sequences represent only 10.2\% of the results.
This demonstrates that multi-digit perception failures are influenced by stopping-policy and sequence-tracking errors rather than purely visual recognition failures. 

We audited 1,600 Level 2 trajectories across models under Textual State and Metric Grid Map containing textual thoughts or traces. It reveals three failure modes. First, agents frequently stop and predict early when they observe one digit and interpret the near-blank space as the end of the sequence, deciding on a sequence before acquiring additional evidence from the other digit. 323 of the predictions in Level 2 just predict one digit. 238 of these cases never explore any pixel of the other digit and 266 expose less than 25\% of the second digit. These episodes also terminate very early by only consuming 15 out of 78 available sensing budget. Second, a discrepancy frequently arises when the agent's Textual State or Metric Grid Map describes exploring a new region, but its actual action revisits previously observed locations. Third, models frequently fail to revise early beliefs: once an incorrect digit identity is stored in the textual state, it is often preserved despite subsequent evidence suggesting a counterhypothesis.

\geminiseven{} rarely stopped or predicted early in this manner. It does not produce any single-digit predictions under the Textual State experiment and produces only 3/100 under the Metric Grid Map. Instead, the dominant failures arise after the agent has represented a state containing two digits. However, it is affected by carrying an incorrect belief about an identified digit. We examined 39 such cases and found that this occurred in 17 cases. Of these 17 cases, 15 retained the incorrect identity in the final prediction. We also find that the model revisits previously explored regions in 45.1\% of moves under the Textual State setting and completely misses exploring one digit in 24/100 episodes. The Grid Map reduces the revisit rate to 13.6\%, but the model still misses exploring one digit in 23/100 episodes. These results indicate that acquiring additional evidence is insufficient when early belief or spatial assignments remain incorrectly encoded in the perceptual state.

\paragraph{The Relationship Between Visual Coverage and Success (\Cref{fig:mnist-pro-multidigit-failure-analysis}c).}
To see whether agents fail from poor exploration or poor memory integration, we compare prediction accuracy against the \emph{minimum observed coverage of either digit}. 
Looking at both digits is necessary for success.
When minimum coverage is $\le 25\%$, exact-sequence accuracy is just $1.0\%$. 
Although accuracy increases as agents explore more ($11.3\%$ for 25--50\% coverage and $17.6\%$ for 50--75\% coverage), its highest is only $26.0\%$ even when agents see more than $75\%$ of both digits. 
Simply looking at the digits is therefore not the main barrier. 
Instead, the failure happens in working memory, where models struggle to organize, store, and interpret their sequential observations.

\subsubsection{External Perceptual-State Construction Helps}
\label{sec:canvas-eval}
The Memory Canvas experiments provide a complementary observation (\Cref{tab:memory_regimes}). In this
condition, observations are programmatically combined into an image canvas
after every exploration step, and the evolving representation is used for
subsequent decisions. For Gemini-3 models, performance consistently improves across Level 1 and Level 2. For \gptterra{} on Level 1, the in-loop/online Memory Canvas
reaches 31.0\% accuracy after 6.91 steps, whereas the Offline Canvas
construction from the Image Only trajectory reaches 49.0\% after 18.19
steps and offline construction from the Textual State trajectory reaches
56.0\% after 24.09 steps. The substantially earlier commitment of the
in-loop Memory Canvas may therefore contribute to its lower accuracy.

The performance improvement is more consistent for the Offline Canvas (\Cref{tab:memory_regimes}) experiment in which canvases are created from primary glimpses produced by the agents under Image Only and Textual State setups. Both these setups, once combined with canvas-based post-processing, significantly improve the performance for every agent. Textual State + Canvas consistently outperforms the Image Only + Canvas setting. 

\claudeopus{} and \claudefable{} acquire substantial visual evidence under the Textual State condition: their stroke coverage is 83.0\%/80.5\% on Level 1 and 80.5\%/82.0\% on Level 2, respectively, although their corresponding accuracies remain 41.0\%/38.0\% and 13.0\%/18.0\%. We therefore replay the same trajectories and programmatically consolidate the observed glimpses into a coordinate-aligned offline canvas, leaving evidence acquisition and stopping unchanged under the Offline Canvas setting. Accuracy rises to 76.0\% and 67.0\% for \claudeopus{} and to 81.0\% and 69.0\% for \claudefable{} on Levels 1 and 2, respectively. Offline canvases built from Image Only trajectories also improve accuracy, but less strongly on Level 2 (36.0\% versus 67.0\% for \claudeopus{}; 42.0\% versus 69.0\% for \claudefable{}). Because the trajectories are fixed, these gains localize much of the error downstream of evidence acquisition, to the organization and/or interpretation of accumulated observations.

\begin{table*}[t]
\centering
\resizebox{\linewidth}{!}{%
\begin{tabular}{lccc|ccc|ccc|ccc|ccc}
\toprule
\multicolumn{16}{c}{\textbf{(a) Accuracy (\%) $\uparrow$}} \\
\midrule
\multirow{2}{*}{\textbf{Model}}
& \multicolumn{3}{c}{\textbf{Image Only}}
& \multicolumn{3}{c}{\textbf{Image Only + Canvas}}
& \multicolumn{3}{c}{\textbf{Textual State}}
& \multicolumn{3}{c}{\textbf{Textual State + Canvas}}
& \multicolumn{3}{c}{\textbf{Memory Canvas}} \\
& \multicolumn{3}{c}{($\hist{=}\infty$)}
& \multicolumn{3}{c}{($\hist{=}\infty$, offline)}
& \multicolumn{3}{c}{($\hist{=}1$)}
& \multicolumn{3}{c}{($\hist{=}1$, offline)}
& \multicolumn{3}{c}{($\hist{=}1$, online)} \\
\cmidrule(lr){2-4} \cmidrule(lr){5-7} \cmidrule(lr){8-10} \cmidrule(lr){11-13} \cmidrule(lr){14-16}
& L1 & L2 & \textbf{Avg} & L1 & L2 & \textbf{Avg} & L1 & L2 & \textbf{Avg} & L1 & L2 & \textbf{Avg} & L1 & L2 & \textbf{Avg} \\
\midrule
\geminipro{}   & 38.0 & \phantom{0}5.0 & 21.5 & 66.0 & 10.0 & 38.0 & 56.0 & 20.0 & 38.0 & 87.0 & 35.0 & \textbf{61.0} & 52.0 & 16.0 & 34.0 \\
\geminisix{}   & 53.0 & 18.0 & 35.5 & 77.0 & 35.0 & 56.0 & 47.0 & 15.0 & 31.0 & 89.0 & 49.0 & \textbf{69.0} & 61.0 & 34.0 & 47.5 \\
\geminiseven{} & 65.0 & 22.0 & 43.5 & 74.0 & 41.0 & 57.5 & 54.0 & \phantom{0}6.0 & 30.0 & 81.0 & 51.0 & \textbf{66.0} & 68.0 & 38.0 & 53.0 \\
\claudeopus{}  & 43.0 & 13.0 & 28.0 & 63.0 & 36.0 & 49.5 & 41.0 & 13.0 & 27.0 & 76.0 & 67.0 & \textbf{71.5} & -- & -- & -- \\
\claudefable{} & 47.0 & 15.0 & 31.0 & 74.0 & 42.0 & 58.0 & 38.0 & 18.0 & 28.0 & 81.0 & 69.0 & \textbf{75.0} & -- & -- & -- \\
\gptterra{}    & 35.0 & \phantom{0}1.0 & 18.0 & 49.0 & \phantom{0}1.0 & 25.0 & 25.0 & \phantom{0}0.0 & 12.5 & 56.0 & \phantom{0}4.0 & \textbf{30.0} & 31.0 & \phantom{0}1.0 & 16.0 \\
\gptsol{}      & 37.0 & \phantom{0}1.0 & 19.0 & 49.0 & \phantom{0}1.0 & 25.0 & 29.0 & \phantom{0}0.0 & 14.5 & 55.0 & \phantom{0}6.0 & \textbf{30.5} & 33.0 & \phantom{0}1.0 & 17.0 \\
\gptsolxhigh{} & 56.0 & 29.0 & 42.5 & 68.0 & 59.0 & \textbf{63.5} & 37.0 & \phantom{0}3.0 & 20.0 & 65.0 & 26.0 & 45.5 & -- & -- & -- \\
\end{tabular}%
}

\vspace{6pt}

\resizebox{0.99\linewidth}{!}{%
\begin{tabular}{lccc|ccc|ccc|ccc|ccc}
\toprule
\multicolumn{16}{c}{\textbf{(b) Average Steps $\downarrow$}} \\
\midrule
\multirow{2}{*}{\textbf{Model}}
& \multicolumn{3}{c}{\textbf{Image Only}}
& \multicolumn{3}{c}{\textbf{Image Only + Canvas}}
& \multicolumn{3}{c}{\textbf{Textual State}}
& \multicolumn{3}{c}{\textbf{Textual State + Canvas}}
& \multicolumn{3}{c}{\textbf{Memory Canvas}} \\
& \multicolumn{3}{c}{($\hist{=}\infty$)}
& \multicolumn{3}{c}{($\hist{=}\infty$, offline)}
& \multicolumn{3}{c}{($\hist{=}1$)}
& \multicolumn{3}{c}{($\hist{=}1$, offline)}
& \multicolumn{3}{c}{($\hist{=}1$, online)} \\
\cmidrule(lr){2-4} \cmidrule(lr){5-7} \cmidrule(lr){8-10} \cmidrule(lr){11-13} \cmidrule(lr){14-16}
& L1 & L2 & \textbf{Avg} & L1 & L2 & \textbf{Avg} & L1 & L2 & \textbf{Avg} & L1 & L2 & \textbf{Avg} & L1 & L2 & \textbf{Avg} \\
\midrule
\geminipro{}   & \phantom{0}4.52 & 11.95 & \phantom{0}8.23 & \phantom{0}4.52 & 11.95 & \phantom{0}8.23 & \phantom{0}7.18 & 18.28 & 12.73 & \phantom{0}7.18 & 18.28 & 12.73 & \phantom{0}4.12 & 16.78 & 10.45 \\
\geminisix{}   & \phantom{0}5.62 & 15.11 & 10.37 & \phantom{0}5.62 & 15.11 & 10.37 & \phantom{0}9.35 & 39.29 & 24.32 & \phantom{0}9.35 & 39.29 & 24.32 & \phantom{0}5.03 & 17.57 & 11.30 \\
\geminiseven{} & \phantom{0}5.26 & 14.69 & \phantom{0}9.97 & \phantom{0}5.26 & 14.69 & \phantom{0}9.97 & \phantom{0}7.92 & 34.84 & 21.38 & \phantom{0}7.92 & 34.84 & 21.38 & \phantom{0}5.56 & 20.40 & 12.98 \\
\claudeopus{}  & \phantom{0}7.43 & 22.50 & 14.97 & \phantom{0}7.43 & 22.50 & 14.97 & 11.61 & 39.90 & 25.76 & 11.61 & 39.90 & 25.76 & -- & -- & -- \\
\claudefable{} & \phantom{0}7.27 & 22.30 & 14.79 & \phantom{0}7.27 & 22.30 & 14.79 & 11.30 & 35.93 & 23.62 & 11.30 & 35.93 & 23.62 & -- & -- & -- \\
\gptterra{}    & 18.19 & 39.67 & 28.93 & 18.19 & 39.67 & 28.93 & 24.09 & 69.67 & 46.88 & 24.09 & 69.67 & 46.88 & \phantom{0}6.91 & 44.56 & 25.73 \\
\gptsol{}      & \phantom{0}6.53 & 16.98 & 11.76 & \phantom{0}6.53 & 16.98 & 11.76 & \phantom{0}7.01 & 22.52 & 14.77 & \phantom{0}7.01 & 22.52 & 14.77 & \phantom{0}4.09 & 14.52 & \phantom{0}9.30 \\
\gptsolxhigh{} & \phantom{0}9.59 & 35.43 & 22.51 & \phantom{0}9.59 & 35.43 & 22.51 & \phantom{0}9.19 & 28.00 & 18.59 & \phantom{0}9.19 & 28.00 & 18.59 & -- & -- & -- \\
\bottomrule
\end{tabular}
}

\caption{Accuracy and average steps across perceptual state construction methods, per level and averaged. Offline conditions replay the trajectory of the live condition to their left, so their step counts in (b) are identical by construction. \textbf{Bold} indicates each model's best average accuracy.}
\label{tab:memory_regimes}
\end{table*}

\begin{table*}[t]
\centering
\small
\resizebox{0.99\linewidth}{!}{%
\begin{tabular}{lcccccc|cccccc}
\toprule
\multirow{3}{*}{\textbf{Condition}}
& \multicolumn{6}{c}{\gptterra{}} & \multicolumn{6}{c}{\geminiseven{}} \\
\cmidrule(lr){2-7} \cmidrule(lr){8-13}
& \multicolumn{3}{c}{\textbf{Acc (\%) $\uparrow$}} & \multicolumn{3}{c}{\textbf{Avg. Steps $\downarrow$}}
& \multicolumn{3}{c}{\textbf{Acc (\%) $\uparrow$}} & \multicolumn{3}{c}{\textbf{Avg. Steps $\downarrow$}} \\
\cmidrule(lr){2-4} \cmidrule(lr){5-7} \cmidrule(lr){8-10} \cmidrule(lr){11-13}
& L1 & L2 & \textbf{Avg} & L1 & L2 & \textbf{Avg} & L1 & L2 & \textbf{Avg} & L1 & L2 & \textbf{Avg} \\
\midrule
\multicolumn{13}{l}{\textit{Non-harness}} \\
Image Only ($\hist{=}\infty$)   & 35.0 & \phantom{0}1.0 & 18.0 & 18.19 & 39.67 & 28.93 & 65.0 & 22.0 & 43.5 & \phantom{0}5.26 & 14.69 & \phantom{0}9.97 \\
Textual State ($\hist{=}1$)     & 25.0 & \phantom{0}0.0 & 12.5 & 24.09 & 69.67 & 46.88 & 54.0 & \phantom{0}6.0 & 30.0 & \phantom{0}7.92 & 34.84 & 21.38 \\
Metric Grid Map ($\hist{=}1$)   & 24.0 & \phantom{0}0.0 & 12.0 & 21.92 & 62.43 & 42.17 & 57.0 & 28.0 & 42.5 & \phantom{0}6.82 & 20.37 & 13.60 \\
Memory Canvas ($\hist{=}1$, online) & 31.0 & \phantom{0}1.0 & 16.0 & \phantom{0}6.91 & 44.56 & 25.73 & 68.0 & 38.0 & 53.0 & \phantom{0}5.56 & 20.40 & 12.98 \\
\midrule
\multicolumn{13}{l}{\textit{Non-harness}} \\
Image Only + Canvas             & 49.0 & \phantom{0}1.0 & 25.0 & 18.19 & 39.67 & 28.93 & 74.0 & 41.0 & 57.5 & \phantom{0}5.26 & 14.69 & \phantom{0}9.97 \\
Textual State + Canvas          & \textbf{56.0} & \phantom{0}4.0 & \textbf{30.0} & 24.09 & 69.67 & 46.88 & \textbf{81.0} & 51.0 & \textbf{66.0} & \phantom{0}7.92 & 34.84 & 21.38 \\
\midrule
\multicolumn{13}{l}{\textit{Agentic harness, shell interface}} \\
Agentic shell                   & 23.0 & \phantom{0}\textbf{6.0} & 14.6 & 14.42 & 40.46 & 27.44 & -- & -- & -- & -- & -- & -- \\
\midrule
\multicolumn{13}{l}{\textit{Agentic harness, native MCP}} \\

Agentic MCP
& 31.0 & -- & 31.0
& 10.24 & -- & 10.24
& 88.0 & 63.0 & 75.5
& 15.67 & 41.00 & 28.33 \\

\quad + persistent memory
& 32.0 & -- & 32.0
& \phantom{0}9.91 & -- & \phantom{0}9.91
& 85.0 & 62.0 & 73.5
& 15.30 & 52.42 & 33.86 \\

\quad + memory + feedback
& 28.0 & -- & 28.0
& 11.16 & -- & 11.16
& 86.0 & 63.0 & 74.5
& 16.02 & 63.67 & 39.84 \\
\bottomrule
\end{tabular}
}
\caption{\gptterra{} and \geminiseven{} on \method{} using harness.}
\label{tab:terra_harness}
\end{table*}

\subsubsection{Autonomous Harnesses Help but Persistent Memory Doesn't}

The results in \Cref{tab:terra_harness} show the performance of harness
systems using \gptterra{} and \geminiseven{}. In this setup, the agents are
automatically invoked by the harness and can perform autonomous agentic
work, including creating, reading, and writing files, writing and executing
code, and discovering strategies for solving the task. Within each episode,
the harness maintains the conversation and associated tool results, thereby
providing access to a richer interaction history than the non-harness
conditions.

\paragraph{\gptterra{}.}
We do not observe a meaningful accuracy improvement for \gptterra{} under
the harness. On Level 1, Agentic MCP achieves 31.0\% accuracy, compared
with 35.0\% for Image Only, 25.0\% for Textual State, and 31.0\% for the
in-loop/online Memory Canvas. The offline Textual State + Canvas condition reaches
56.0\%. This indicates that greater tool autonomy alone is insufficient for
\gptterra{} to overcome its perceptual-state construction limitations.
Agentic MCP uses 10.24 steps, compared with 18.19 for Image Only and 24.09
for Textual State, although the in-loop/online Memory Canvas terminates earlier,
after 6.91 steps.

In the persistent-memory setup, agents can write information that survives
across otherwise independent episodes. The objective is to test whether
agents can discover, retain, and reuse procedural knowledge. Our trajectory
analysis shows that \gptterra{} identified a potentially useful procedure
for constructing a visual canvas and representing it using ASCII
characters. This procedure remained in its persistent memory across
multiple episodes, but \gptterra{} executed it in only approximately 2\%
of episodes. This reveals a gap between discovering and storing useful
procedural knowledge and invoking it in subsequent episodes. Consistent
with this observation, persistent memory does not materially improve
Level 1 accuracy: Agentic MCP obtains 31.0\%, persistent memory obtains
32.0\%, and memory with correctness feedback obtains 28.0\%. Average
sensing effort also changes only modestly, from 10.24 steps without
persistent memory to 9.91 with persistent memory and 11.16 with memory and
feedback.

\paragraph{\geminiseven{}.}
The behavior of \geminiseven{} is substantially different~\footnote{\claudeopus{} exhibits a similar pattern. In a limited matched evaluation of 26 valid Level~1 episodes, the Claude Code harness achieves 80.8\% accuracy (21/26), compared with 46.2\% (12/26) under native multi-turn interaction, an improvement of 34.6\%. Trace analysis shows that the agent programmatically integrated multiple glimpses in 18/26 episodes and successfully reopened a resulting visual composite in 16/26 episodes. The harness used 22.2 observations per episode on average, compared with 12.3 under native multi-turn interaction ($1.80\times$ as many).}. Compared with
the strongest in-loop/online non-harness configuration, Memory Canvas, Agentic MCP
improves accuracy from 68.0\% to 88.0\% on Level 1 and from 38.0\% to
63.0\% on Level 2. These correspond to gains of 20 and 25 percentage
points, respectively. This improvement is accompanied by substantially
greater sensing effort. On Level 1, Agentic MCP uses 15.67 steps, compared
with 5.56 for Memory Canvas, or approximately $2.8\times$ as many steps.
On Level 2, it uses 41.00 steps, compared with 20.40 for Memory Canvas, or
approximately $2.0\times$ as many. Thus, unlike the non-harness
configuration, the autonomous agent gathers substantially more evidence
before committing to a prediction.

Trajectory inspection shows that \geminiseven{} frequently performs
explicit perceptual-state construction before producing its final answer.
It writes programs such as \texttt{stitch.py} and
\texttt{render\_mnist.py}, reads the observations produced during
exploration, combines their revealed pixels, and renders the resulting
spatial representation as an ASCII or downsampled canvas. The model may
execute this reconstruction repeatedly as new observations become
available, although the frequency varies across episodes. This behavior is
autonomously discovered: the harness permits file and tool use but does not
explicitly prescribe spatial stitching. While without persistent memory, 75\% of samples in Level 1 used perceptual-state construction by assembling the observations into a canvas, the persistent memory counterparts on Level 1 and Level 2 and even without persistent memory on Level 2, used this technique 100\% of the time, i.e., for every sample in the dataset. 

Persistent memory makes this reconstruction strategy more standardized.
Descriptions of the stitching procedure, common filenames, and decoding
heuristics are retained in memory and reused in later episodes. The
executable workspace remains episode-specific, so the agent must recreate
the helper programs from the procedural information stored in memory.
Thus, persistent memory transfers procedural instructions rather than a
persistent executable artifact.

Despite this greater procedural consistency, persistent memory does not
improve \geminiseven{}'s accuracy. Without persistent memory, Agentic MCP
obtains 88.0\% on Level 1 and 63.0\% on Level 2, for an average of 75.5\%.
Adding persistent memory results in 85.0\% and 62.0\%, respectively, for
an average of 73.5\%. Adding correctness feedback produces 86.0\% and
63.0\%, for an average of 74.5\%. Therefore, neither persistent memory nor
correctness feedback exceeds the no-memory Agentic MCP condition.

The effect on sensing effort differs across levels. On Level 1, the average
number of steps remains relatively stable: 15.67 without persistent
memory, 15.30 with persistent memory, and 16.02 with memory and feedback.
On Level 2, however, sensing effort increases substantially from 41.00
steps without persistent memory to 52.42 with persistent memory and 63.67
with memory and feedback. Averaged across levels, the corresponding values
are 28.33, 33.86, and 39.84 steps. Persistent procedural memory therefore
induces longer and more elaborate exploration, particularly on Level 2,
without producing a corresponding improvement in accuracy.

Memory Canvas, which creates a canvas from the observations at every step to aid the model's prediction, performs worse than Image Only + Canvas and Textual State + Canvas, which create the canvas only after completing an initial exploration. This can be attributed to stopping exploration too early or being misguided in the earlier steps, leading to incorrect prediction. By contrast, \geminiseven{} autonomously determines when to construct and
inspect a consolidated representation. This suggests that effective
perceptual-state construction depends not only on which representation is
available, but also on when it is constructed and how its use interacts
with the agent's exploration policy.

The autonomous \geminiseven{} harness effectively acquires observations and organizes them into a coherent sequence when solving Level 2. Its position-wise accuracy is 85\% for the first digit and 73\% for the second digit. In 92/100 episodes, the agent observes at least 75\% of both digits, yet 32 of these episodes are predicted incorrectly. Similarly, 67/100 episodes reveal at least 90\% of both digits, yet 23 of these episodes are predicted incorrectly. Therefore, the harness largely resolves evidence acquisition and the organization of observations into the correct sequence. The remaining bottleneck lies in interpreting the constructed perceptual state.

On Level 2, the \geminiseven{} harness generally starts by reconstructing the digit containing the initial randomly given glimpse. It then crosses to the other digit, attempts to reconstruct it, and occasionally returns for verification. We observe that the first evidence from the second digit is obtained at a median of 17 observations; by that time, the first digit has already reached a median visible coverage of 94.1\%. The midpoint of the canvas is crossed once in 57\% of episodes and twice in 21\% of episodes, while immediate direction reversals during exploration occur in only 2.8\% of effective consecutive moves. Therefore, the improved performance does not arise from exhaustive search, but rather from a more systematic and well-structured process of evidence acquisition and external state construction.

\paragraph{Agents Behave Differently in the Harness Setup.}

Under the native multi-turn setup, \geminiseven{} makes decisions after 8.23 and 28.45 sensing steps on Levels~1 and~2, respectively, achieving 75\% and 47\% accuracy. With the harness, it increases its sensing effort to 15.67 and 41.00 steps and improves its accuracy to 88\% and 63\%. Thus, when \geminiseven{} can externalize, reconstruct, and inspect its perceptual state, it acquires more evidence before deciding. In contrast, under the native multi-turn setup, it exhibits more overoptimistic stopping and makes decisions much earlier. Part of this difference may also arise from system-prompt-based steering: harnesses are optimized for verifiability, and their customized system prompts may encourage agents to spend more time solving the task to achieve better results. This possibility becomes clearer in the Memory Canvas experiment, where we programmatically assemble the observed glimpses into a canvas and provide it to the agent at every step. Yet even in this setting, agents still make incorrect decisions after fewer steps than under the native multi-turn setup.

\subsubsection{Exploration and Spatial Integration}

\begin{table*}[t]
\centering
\small
\resizebox{0.99\linewidth}{!}{%
\begin{tabular}{lcccccccccccccccccc}
\toprule
\textbf{Model} & \multicolumn{6}{c}{\textbf{Image Only ($\hist=\infty$)}} & \multicolumn{6}{c}{\textbf{Textual State ($\hist=1$)}} & \multicolumn{6}{c}{\textbf{Metric Grid Map ($\hist=1$)}} \\
\cmidrule(r){2-7} \cmidrule(r){8-13} \cmidrule(l){14-19}
& \multicolumn{3}{c}{\textbf{Level 1}} & \multicolumn{3}{c}{\textbf{Level 2}} & \multicolumn{3}{c}{\textbf{Level 1}} & \multicolumn{3}{c}{\textbf{Level 2}} & \multicolumn{3}{c}{\textbf{Level 1}} & \multicolumn{3}{c}{\textbf{Level 2}} \\
\cmidrule(lr){2-4} \cmidrule(lr){5-7} \cmidrule(lr){8-10} \cmidrule(lr){11-13} \cmidrule(lr){14-16} \cmidrule(lr){17-19}
& \textbf{$M_T$} & \textbf{$RR_T$ (\%) $\downarrow$} & \textbf{$C_T$ (\%) $\uparrow$} & \textbf{$M_T$} & \textbf{$RR_T$ (\%) $\downarrow$} & \textbf{$C_T$ (\%) $\uparrow$} & \textbf{$M_T$} & \textbf{$RR_T$ (\%) $\downarrow$} & \textbf{$C_T$ (\%) $\uparrow$} & \textbf{$M_T$} & \textbf{$RR_T$ (\%) $\downarrow$} & \textbf{$C_T$ (\%) $\uparrow$} & \textbf{$M_T$} & \textbf{$RR_T$ (\%) $\downarrow$} & \textbf{$C_T$ (\%) $\uparrow$} & \textbf{$M_T$} & \textbf{$RR_T$ (\%) $\downarrow$} & \textbf{$C_T$ (\%) $\uparrow$} \\
\midrule
\multicolumn{19}{l}{\textit{Proprietary Models}} \\
\midrule
\geminipro{} & 3.5 & 3.7 & 57.7 & 10.9 & \textbf{10.0} & 40.4 & 6.2 & 5.3 & \textbf{73.8} & 17.3 & 13.6 & \textbf{55.9} & 6.2 & \textbf{3.6} & 71.7 & 18.2 & 20.2 & 52.9 \\
\geminisix{} & 4.6 & \textbf{5.6} & 65.9 & 14.1 & \textbf{10.6} & 49.0 & 8.3 & 16.3 & \textbf{77.9} & 38.3 & 45.0 & \textbf{65.4} & 6.9 & 6.1 & 71.0 & 19.4 & 13.3 & 56.4 \\
\geminiseven{} & 4.3 & \textbf{4.2} & 66.8 & 13.7 & \textbf{6.3} & 53.9 & 6.9 & 15.8 & \textbf{74.4} & 33.8 & 45.1 & 61.0 & 5.8 & 8.1 & 69.3 & 19.4 & 13.6 & \textbf{61.7} \\
\claudesonnet{} & 16.4 & 46.7 & 71.1 & 44.1 & 54.7 & 45.1 & 13.8 & 34.9 & 72.2 & 74.6 & 64.2 & 50.6 & 12.1 & \textbf{20.2} & \textbf{72.9} & 49.4 & \textbf{46.6} & \textbf{56.4} \\
\claudeopus{} & 6.4 & 9.6 & 72.9 & 21.5 & 20.2 & 61.7 & 10.6 & 18.8 & \textbf{83.0} & 38.9 & 25.4 & \textbf{80.5} & 9.6 & \textbf{8.6} & 81.5 & 28.6 & \textbf{13.6} & 75.4 \\
\claudefable{} & 6.3 & 7.3 & 74.7 & 21.3 & 12.6 & 68.7 & 10.3 & 17.0 & \textbf{80.5} & 34.9 & 19.7 & \textbf{82.0} & 8.9 & \textbf{6.5} & 76.8 & 27.6 & \textbf{9.7} & 76.6 \\
\gptterra{} & 17.2 & 35.8 & 71.0 & 38.7 & 49.3 & 41.5 & 23.1 & 44.4 & 78.1 & 68.7 & 71.6 & 44.4 & 20.9 & \textbf{26.2} & \textbf{81.3} & 61.4 & \textbf{46.1} & \textbf{55.6} \\
\gptsol{} & 5.5 & 12.5 & 58.5 & 16.0 & 26.3 & 35.4 & 6.0 & 20.3 & 60.7 & 21.5 & 33.0 & \textbf{42.5} & 6.9 & \textbf{8.1} & \textbf{64.0} & 20.3 & \textbf{22.2} & 41.1 \\
\gptsolxhigh{} & 8.6 & \textbf{9.9} & 75.3 & 34.4 & \textbf{20.7} & \textbf{75.8} & 8.2 & 13.9 & \textbf{75.3} & 27.0 & 25.8 & 55.5 & 8.8 & 8.6 & 71.8 & 35.1 & 24.9 & 58.4 \\
\midrule
\multicolumn{19}{l}{\textit{Open-Source Models}} \\
\midrule
\qwen{} & 3.8 & 22.1 & 48.1 & 8.2 & 39.7 & 25.4 & 6.0 & 34.2 & 53.0 & 15.9 & 53.6 & 32.1 & 5.9 & \textbf{16.4} & \textbf{56.2} & 12.9 & \textbf{29.7} & \textbf{35.6} \\
\glm{} & 2.5 & \textbf{20.1} & 36.5 & 5.5 & \textbf{17.2} & 20.7 & 19.2 & 61.8 & \textbf{49.4} & 39.1 & 62.3 & \textbf{36.3} & 10.4 & 39.0 & 46.7 & 26.7 & 55.2 & 33.1 \\
\bottomrule
\end{tabular}
}
\caption{Exploration and loop-closure metrics: Average Moves ($M_T$), Revisit Rate ($RR_T \downarrow$), and Stroke Coverage ($C_T \uparrow$) for Level 1 (Single-Digit) and Level 2 (Multi-Digit) tasks across three observation configurations. Bold values indicate the lowest revisit rate or highest stroke coverage within the corresponding model and task configuration.}
\label{tab:revisit_analysis}
\end{table*}

Previous analyses evaluate the high-level impact of state representations and harnesses on final task accuracy, we now turn to the movement mechanics of the agents.
We investigate how different state representations shape exploration behaviors and redundant paths by analyzing two metrics \textbf{Revisit Rate} $RR_T$ and \textbf{Stroke Coverage} $C_T$ (\Cref{tab:revisit_analysis}).

We represent the canvas as a binarized grid $I$ of size $W' \times H$ where stroke pixels have value 0. The set of stroke pixels $S$ is:
\begin{equation}
    S = \{ (u, v) \in \mathbb{N}^2 \mid I(u, v) = 0, \, 0 \le u < W', \, 0 \le v < H \}
\end{equation}

At step $t$, the agent's $h \times w$ glimpse window $B(x_t, y_t)$ starts at top-left coordinate $(x_t, y_t)$:
\begin{equation}
    B(x_t, y_t) = \{ (u, v) \in \mathbb{N}^2 \mid x_t \le u < x_t + w, \, y_t \le v < y_t + h \}
\end{equation}
The explored region $V_T$ is the union of all glimpses visited up to step $T$:
\begin{equation}
    V_T = \bigcup_{t=0}^T B(x_t, y_t)
\end{equation}

Stroke coverage $C_T$ is the fraction of stroke pixels $S$ that lie within $V_T$:
\begin{equation}
    C_T = \frac{|S \cap V_T|}{|S|} = \frac{\left| S \cap \left( \bigcup_{t=0}^T B(x_t, y_t) \right) \right|}{|S|}
\end{equation}

The revisit rate $RR_T$ measures movement redundancy. We count the steps $R_T$ where the agent returns to a previously visited coordinate:
\begin{equation}
    R_T = \sum_{t=1}^T \mathbb{I}\Big( (x_t, y_t) \in \{(x_k, y_k)\}_{k=0}^{t-1} \Big)
\end{equation}
where $\mathbb{I}(\cdot)$ is the indicator function. The rate $RR_T$ is the ratio of revisits to the total movement actions $M_T$:
\begin{equation}
    RR_T = \frac{R_T}{M_T}
\end{equation}

\paragraph{Structured Spatial Maps Reduce Redundant Exploration.}
Unstructured textual states under a zero-lookback visual history ($\hist=1$) do not prevent repetitive circular loops, but structured metric grid maps reduce revisits.
In Level 1, \geminisix{}'s revisit rate falls from 16.3\% with a textual state to 6.1\% with a metric grid map.
In Level 2, \geminiseven{}'s revisit rate drops from 45.1\% to 13.6\% (qualitative examples of Level 1 and Level 2 trajectories are shown in \Cref{fig:L1,fig:L2}, respectively).
These results show that spatial memory needs structured representations to prevent tracking drift under zero-lookback visual history.

\paragraph{Unbounded Visual History Terminates Exploration Prematurely.}
Agents that rely only on visual history ($\hist=\infty$) often terminate search early, which results in lower canvas coverage than textual memory configurations with a constrained horizon ($\hist=1$).
In Level 1, \geminipro{} achieves only 57.7\% coverage across 3.5 moves with full visual history, but reaches 71.7\% coverage across 6.2 moves with a zero-lookback metric grid map.
This pattern holds for \gptterra{}, whose coverage increases from 71.0\% to 81.3\% when it uses the metric grid map instead of full visual history.

\paragraph{Task Complexity Exposes Textual Memory Tracking Drift.}
The transition from single-digit (Level 1) to multi-digit (Level 2) shows tracking degradation in unstructured memory, whereas structured spatial memory remains stable as task complexity increases.
In Level 2, \geminiseven{}'s textual state revisit rate reaches 45.1\%, compared to 15.8\% in Level 1.
The metric grid map keeps the Level 2 revisit rate at 13.6\% and preserves 61.7\% canvas coverage.
Structured spatial coordinates keep tracking consistent as the task environment grows in scale and complexity.

\paragraph{Open-Source Limitations in Spatial State Construction.}
Open source models struggle with perceptual state construction and exploration efficiency compared to proprietary models.
With a textual state, \glm{} falls into severe loops, which leads to a 61.8\% revisit rate in Level 1 and 62.3\% in Level 2.
Even with a metric grid map, \glm{} and \qwen{} struggle to search the canvas, and \glm{} covers only 46.7\% in Level 1 and 33.1\% in Level 2.
These lower metrics suggest that open-source models struggle with path integration and systematic planning under partial observability over long trajectories.

\paragraph{Distinct Behavioral Tradeoffs in Coverage and Efficiency.}
Proprietary models show distinct tradeoffs between exploration coverage and path efficiency.
\claudefable{} consistently conducts exhaustive search, having the highest stroke coverage across Level 1 (80.5\% Textual, 76.8\% Grid Map) and Level 2 (82.0\% Textual, 76.6\% Grid Map) while keeping revisit rates low.
Gemini-3 models tend to terminate search early, taking fewer steps on average but having lower overall stroke coverage.
\gptterra{} shows highly redundant exploration trajectories, taking 61.4 moves and having a 46.1\% revisit rate under the Level 2 metric grid map configuration.

\subsubsection{Digit Topology}

\begin{table*}[t]
\centering
\small
\resizebox{0.99\linewidth}{!}{%
\begin{tabular}{lccccccccc}
\toprule
\textbf{Model} & \multicolumn{3}{c}{\textbf{Image Only Baseline ($\hist=\infty$)}} & \multicolumn{3}{c}{\textbf{Textual State ($\hist=1$)}} & \multicolumn{3}{c}{\textbf{Metric Grid Map ($\hist=1$)}} \\
\cmidrule(r){2-4}
\cmidrule(r){5-7}
\cmidrule(r){8-10}
& \textbf{Strokes (\%) $\uparrow$} & \textbf{Curves (\%) $\uparrow$} & \textbf{Loops (\%) $\uparrow$} & \textbf{Strokes (\%) $\uparrow$} & \textbf{Curves (\%) $\uparrow$} & \textbf{Loops (\%) $\uparrow$} & \textbf{Strokes (\%) $\uparrow$} & \textbf{Curves (\%) $\uparrow$} & \textbf{Loops (\%) $\uparrow$} \\
\midrule
\multicolumn{10}{l}{\textit{Proprietary Models}} \\
\midrule
\geminipro{} & 43.3 & \textbf{56.7} & 20.0 & \textbf{66.7} & 53.3 & \textbf{50.0} & 56.7 & \textbf{56.7} & 47.5 \\
\geminisix{} & 60.0 & \textbf{50.0} & \textbf{50.0} & \textbf{63.3} & 30.0 & 47.5 & \textbf{63.3} & 36.7 & 47.5 \\
\geminiseven{} & \textbf{76.7} & 53.3 & \textbf{65.0} & 66.7 & 36.7 & 57.5 & 73.3 & \textbf{56.7} & 45.0 \\
\claudesonnet{} & \textbf{50.0} & 13.3 & 12.5 & 23.3 & 20.0 & \textbf{22.5} & 23.3 & \textbf{23.3} & 12.5 \\
\claudeopus{} & 56.7 & \textbf{43.3} & 32.5 & \textbf{60.0} & 23.3 & \textbf{40.0} & 43.3 & 33.3 & 32.5 \\
\claudefable{} & \textbf{66.7} & \textbf{26.7} & 47.5 & 60.0 & 23.3 & 32.5 & 56.7 & \textbf{26.7} & \textbf{50.0} \\
\gptterra{} & \textbf{50.0} & \textbf{40.0} & \textbf{20.0} & 30.0 & 26.7 & \textbf{20.0} & 26.7 & 26.7 & \textbf{20.0} \\
\gptsol{} & \textbf{50.0} & \textbf{30.0} & \textbf{32.5} & 46.7 & 26.7 & 17.5 & \textbf{50.0} & 20.0 & 27.5 \\
\gptsolxhigh{} & \textbf{76.7} & \textbf{40.0} & \textbf{52.5} & 50.0 & 30.0 & 32.5 & 53.3 & 33.3 & 32.5 \\
\midrule
\multicolumn{10}{l}{\textit{Open-Source Models}} \\
\midrule
\qwen{} & 36.7 & \textbf{33.3} & 17.5 & 43.3 & 26.7 & \textbf{25.0} & \textbf{56.7} & \textbf{33.3} & 22.5 \\
\glm{} & 33.3 & 6.7 & 5.0 & \textbf{36.7} & \textbf{20.0} & 7.5 & 30.0 & \textbf{20.0} & \textbf{12.5} \\
\bottomrule
\end{tabular}
}
\caption{Task success rates by topological complexity across the three configurations. Digits are grouped into simple strokes (1, 4, 7), high-curvature curves (2, 3, 5), and closed loops (0, 6, 8, 9). \textbf{Bold} values show the best performance for each model within each complexity category across the three configurations.}
\label{tab:digit_fine_grained}
\end{table*}

Beyond spatial efficiency, the visual geometry of digits may also affect the agent's spatial integration.
We analyze how different topological structures influence the tracking and classification success.
\Cref{tab:digit_fine_grained} reports the success rates for three categories: geometric strokes (digits 1, 4, and 7), curved digits (digits 2, 3, and 5), and closed loops (digits 0, 6, 8, and 9). 

\paragraph{Topological Complexity Affects Performance.}
Classification success follows a topological hierarchy.
Simple strokes like 1, 4, and 7 are the easiest to track and recognize.
Curved digits like 2, 3, and 5 are harder, and closed loops like 0, 6, 8, and 9 are the most difficult.
Across nearly all models, success rates drop as topological complexity increases.

\paragraph{Metric Grid Maps Improve Curve Tracing.}
Maintaining a metric coordinate map improves performance on high-curvature digits.
Under the zero-lookback constraint ($\hist=1$), \geminiseven{}'s success rate on curves increases from 36.7\% with a textual state to 56.7\% with a metric grid map.
Similarly, curve accuracy for \claudeopus{} improves from 23.3\% to 33.3\%, and \qwen{} improves from 26.7\% to 33.3\%.

\paragraph{Performance Drops on Closed Loops with Metric Grid Maps.}
The metric grid map builds directly on the textual state, giving the model access to both representations. 
Adding this spatial map, however, lowers accuracy on digits with closed loops.
Under the zero-lookback constraint ($\hist=1$), \geminiseven{}'s loop accuracy drops from 57.5\% with only the textual state to 45.0\% with the metric grid map.
Loop accuracy also decreases for \claudeopus{} (from 40.0\% to 32.5\%) and \claudesonnet{} (from 22.5\% to 12.5\%).
While the spatial map helps with curve-tracing, these consistent drops suggest that it introduces spatial noise or confusion, leading to worse performance.

\subsection{Stress Tests}

Here, we vary two configurations: the length of the glimpse history and the glimpse window size.

\subsubsection{Visual-History Horizon}

\begin{table*}[t]
\centering
\small
\resizebox{0.99\linewidth}{!}{%
\begin{tabular}{lccccccccc}
\toprule
\multirow{3}{*}{\textbf{Model}}
& \multirow{3}{*}{\textbf{History ($\hist$)}}
& \multicolumn{4}{c}{\textbf{Level 1: Single-Digit}}
& \multicolumn{4}{c}{\textbf{Level 2: Multi-Digit}} \\
\cmidrule(lr){3-6}
\cmidrule(lr){7-10}
& & \multicolumn{2}{c}{\textbf{Textual State}}
& \multicolumn{2}{c}{\textbf{Metric Grid Map}}
& \multicolumn{2}{c}{\textbf{Textual State}}
& \multicolumn{2}{c}{\textbf{Metric Grid Map}} \\
\cmidrule(lr){3-4}
\cmidrule(lr){5-6}
\cmidrule(lr){7-8}
\cmidrule(lr){9-10}
& & \textbf{Acc (\%) $\uparrow$} & \textbf{Avg. Steps $\downarrow$} & \textbf{Acc (\%) $\uparrow$} & \textbf{Avg. Steps $\downarrow$}
& \textbf{Acc (\%) $\uparrow$} & \textbf{Avg. Steps $\downarrow$} & \textbf{Acc (\%) $\uparrow$} & \textbf{Avg. Steps $\downarrow$} \\
\midrule
\multirow{4}{*}{\geminipro{}} & 1       & 56.0 & 7.18 & 53.0 & 7.15 & \textbf{20.0} & 18.28 & 14.0 & 19.19 \\
                        & 2       & 48.0 & 5.99 & 53.0 & 6.75 & 16.0 & 19.29 & 14.0 & 17.91 \\
                        & 4       & 50.0 & 5.89 & 49.0 & \textbf{6.50} & 15.0 & 16.65 & 16.0 & \textbf{16.83} \\
                        & $\infty$ & \textbf{57.0} & \textbf{5.48} & \textbf{57.0} & 6.56 & 16.0 & \textbf{16.40} & \textbf{22.0} & 17.06 \\
\midrule
\multirow{4}{*}{\geminisix{}}       & 1       & 47.0 & 9.35 & 49.0 & 7.92 & \textbf{15.0} & 39.29 & 10.0 & 20.44 \\
                        & 2       & 52.0 & 7.33 & 45.0 & 7.09 & 12.0 & 32.39 & \textbf{18.0} & 18.83 \\
                        & 4       & 53.0 & 5.97 & \textbf{54.0} & 6.06 & \textbf{15.0} & 23.80 & 14.0 & 17.07 \\
                        & $\infty$ & \textbf{58.0} & \textbf{5.45} & 51.0 & \textbf{5.82} & 13.0 & \textbf{15.89} & 13.0 & \textbf{16.53} \\
\midrule
\multirow{4}{*}{\geminiseven{}}       & 1       & 54.0 & 7.92 & 57.0 & 6.82 & 6.0 & 34.84 & 28.0 & 20.37 \\
                        & 2       & 60.0 & 6.80 & 63.0 & 6.17 & 10.0 & 30.68 & 23.0 & 19.96 \\
                        & 4       & 62.0 & 5.85 & 60.0 & 5.63 & 16.0 & 26.65 & 20.0 & 19.56 \\
                        & $\infty$ & \textbf{64.0} & \textbf{5.44} & \textbf{64.0} & \textbf{5.49} & \textbf{23.0} & \textbf{15.90} & \textbf{35.0} & \textbf{17.68} \\
\bottomrule
\end{tabular}%
}
\caption{Ablation results for varying glimpse history $\hist \in \{1, 2, 4, \infty\}$ on Level 1 (Single-Digit) and Level 2 (Multi-Digit) tasks under both the Textual State and Metric Grid Map configurations. Bold values highlight the best performance (highest accuracy $\uparrow$ or lowest average steps $\downarrow$) within each model block across the different history lengths.}
\label{tab:horizon_ablation}
\end{table*}

\Cref{tab:horizon_ablation} shows the results across different visual glimpse history lengths, with $\hist$ varied from 1 to $\infty$.
For $\hist=1$, the agent has zero visual lookback and relies on its current frame and memory.
Intermediate horizon of $\hist=2$ and $\hist=4$ retain the last few visual frames in the context window, and $\hist=\infty$ allows full lookback.

\paragraph{Search Efficiency Gains With Longer Glimpse History.}
Increasing the glimpse history $\hist$ reduces the exploration steps across all models and tasks.
For example, in Level 1 tasks with the Textual State representation, increasing the history from zero lookback ($\hist=1$) to unbounded history ($\hist=\infty$) reduces the average number of steps required. 
Specifically, the average steps decrease from 7.18 to 5.48 for \geminipro{}, from 9.35 to 5.45 for \geminisix{}, and from 7.92 to 5.44 for \geminiseven{}.
The benefit is larger on Level 2 tasks, where agents must plan over longer horizons.
For the Level 2 Textual State representation, \geminisix{}'s average steps drop by nearly 60\%, going from 39.29 ($\hist=1$) to 15.89 ($\hist=\infty$).
Similarly, \geminiseven{}'s average steps drop from 34.84 to 15.90 under the same conditions.

\paragraph{Longer Glimpse Histories Improve Task Success.}
On Level 1 tasks, extending the visual lookback from 1 to $\infty$ improves accuracy.
For instance, with the Textual State representation, \geminiseven{}'s success rate increases from 54.0\% to 64.0\%, while \geminisix{}'s accuracy increases from 47.0\% to 58.0\%.
The improvement is also shown in the Metric Grid Map representation, where \geminiseven{}'s accuracy increases from 57.0\% to 64.0\%.
The effect of history length is more pronounced on Level 2 tasks, which require tracking and ordering multiple digits.
In Level 2 tasks with the Textual State representation, \geminiseven{}'s accuracy is 6.0\% at $\hist=1$, and increases to 23.0\% at $\hist=\infty$.
With the Metric Grid Map, its accuracy increases from 28.0\% to 35.0\%.

\subsubsection{Glimpse Size}
\Cref{tab:channel_capacity} reports the results of varying the glimpse window size during the agent's exploration. We observe a highly consistent trend: performance improves while the average number of steps decreases as the window size increases. For example, \geminiseven{}'s accuracy improves on Level 2 Image Only multi-digit recognition from 3\% to 74\% as we increase the window size from $32 \times 32$ to $128 \times 128$. Larger glimpse windows reveal more visual evidence at each observation, reducing the amount of exploration required and substantially improving recognition accuracy.

\begin{table*}[t]
\centering
\scriptsize
\resizebox{0.99\linewidth}{!}{%
\begin{tabular}{ll|cccccc|cccccc|cccccc}
\toprule
\multirow{3}{*}{\textbf{Model}}
& \multirow{3}{*}{\textbf{Window}}
& \multicolumn{6}{c}{\textbf{Image Only Baseline} ($\hist{=}\infty$)}
& \multicolumn{6}{c}{\textbf{Textual State} ($\hist{=}1$)}
& \multicolumn{6}{c}{\textbf{Metric Grid Map} ($\hist{=}1$)} \\
\cmidrule(lr){3-8} \cmidrule(lr){9-14} \cmidrule(lr){15-20}
& 
& \multicolumn{3}{c}{\textbf{Acc. (\%) $\uparrow$}} & \multicolumn{3}{c}{\textbf{Steps $\downarrow$}}
& \multicolumn{3}{c}{\textbf{Acc. (\%) $\uparrow$}} & \multicolumn{3}{c}{\textbf{Steps $\downarrow$}}
& \multicolumn{3}{c}{\textbf{Acc. (\%) $\uparrow$}} & \multicolumn{3}{c}{\textbf{Steps $\downarrow$}} \\
\cmidrule(lr){3-5} \cmidrule(lr){6-8} \cmidrule(lr){9-11} \cmidrule(lr){12-14} \cmidrule(lr){15-17} \cmidrule(lr){18-20}
& & L1 & L2 & \textbf{Avg} & L1 & L2 & \textbf{Avg} & L1 & L2 & \textbf{Avg} & L1 & L2 & \textbf{Avg} & L1 & L2 & \textbf{Avg} & L1 & L2 & \textbf{Avg} \\
\midrule
\multirow{3}{*}{\geminipro{}} & $128 \times 128$ & 88.0 & 61.0 & 74.5 & 2.48 & 6.53 & 4.50 & 79.0 & 62.0 & 70.5 & 3.41 & 8.35 & 5.88 & 82.0 & 51.0 & 66.5 & 3.19 & 7.43 & 5.31 \\
 & $64 \times 64$ & 38.0 & \phantom{0}5.0 & 21.5 & 4.52 & 11.95 & 8.23 & 56.0 & 20.0 & 38.0 & 7.18 & 18.28 & 12.73 & 53.0 & 14.0 & 33.5 & 7.15 & 19.19 & 13.17 \\
 & $32 \times 32$ & 30.0 & \phantom{0}0.0 & 15.0 & 8.41 & 19.75 & 14.08 & 22.0 & \phantom{0}1.0 & 11.5 & 12.92 & 35.76 & 24.34 & 21.0 & \phantom{0}4.0 & 12.5 & 12.65 & 35.85 & 24.25 \\
\midrule
\multirow{3}{*}{\geminisix{}} & $128 \times 128$ & 85.0 & 61.0 & 73.0 & 2.48 & 6.39 & 4.43 & 78.0 & 61.0 & 69.5 & 3.26 & 12.06 & 7.66 & 75.0 & 41.0 & 58.0 & 3.05 & 9.99 & 6.52 \\
 & $64 \times 64$ & 53.0 & 18.0 & 35.5 & 5.62 & 15.11 & 10.37 & 47.0 & 15.0 & 31.0 & 9.35 & 39.29 & 24.32 & 49.0 & 10.0 & 29.5 & 7.92 & 20.44 & 14.18 \\
 & $32 \times 32$ & 24.0 & \phantom{0}2.0 & 13.0 & 11.50 & 30.41 & 20.95 & 19.0 & \phantom{0}3.0 & 11.0 & 17.70 & 55.75 & 36.73 & 21.0 & \phantom{0}3.0 & 12.0 & 14.25 & 30.73 & 22.49 \\
\midrule
\multirow{3}{*}{\geminiseven{}} & $128 \times 128$ & 89.0 & 74.0 & 81.5 & 2.48 & 6.02 & 4.25 & 83.0 & 60.0 & 71.5 & 2.99 & 11.14 & 7.07 & 84.0 & 67.0 & 75.5 & 2.64 & 7.84 & 5.24 \\
 & $64 \times 64$ & 65.0 & 22.0 & 43.5 & 5.26 & 14.69 & 9.97 & 54.0 & \phantom{0}6.0 & 30.0 & 7.92 & 34.84 & 21.38 & 57.0 & 28.0 & 42.5 & 6.82 & 20.37 & 13.60 \\
 & $32 \times 32$ & 30.0 & \phantom{0}3.0 & 16.5 & 12.80 & 33.35 & 23.08 & 28.0 & \phantom{0}2.0 & 15.0 & 16.16 & 49.98 & 33.07 & 30.0 & 10.0 & 20.0 & 13.52 & 40.40 & 26.96 \\
\bottomrule
\end{tabular}
}
\caption{Glimpse channel capacity ablation study results under varying glimpse window sizes ($w \times w$) on Single-Digit (Level 1) and Multi-Digit (Level 2) tasks. Results are evaluated under both zero-visual-lookback ($\hist=1$) and unbounded history ($\hist=\infty$).}
\label{tab:channel_capacity}
\end{table*}

\section{TL;DR: 1) Seen $\neq$ Using; 2) Same Evidence + Better Representation $\rightarrow$ Better Use}

\begin{figure*}[t]
    \centering

    \begin{subfigure}[t]{0.49\textwidth}
        \centering
        \includegraphics[width=\linewidth]{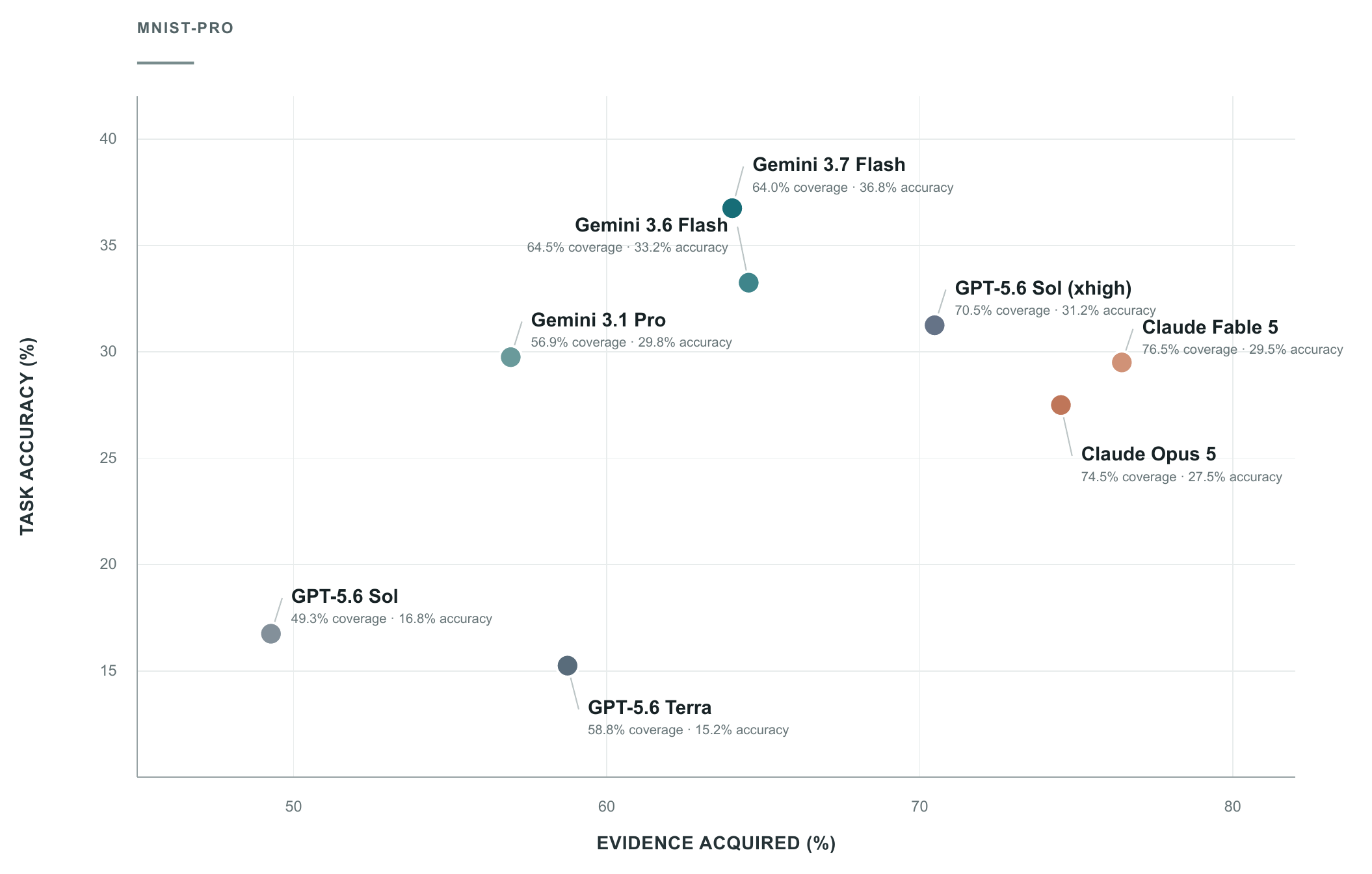}
        \caption{Evidence acquisition versus native task accuracy.}
        \label{fig:evidence-native}
    \end{subfigure}
    \hfill
    \begin{subfigure}[t]{0.49\textwidth}
        \centering
        \includegraphics[width=\linewidth]{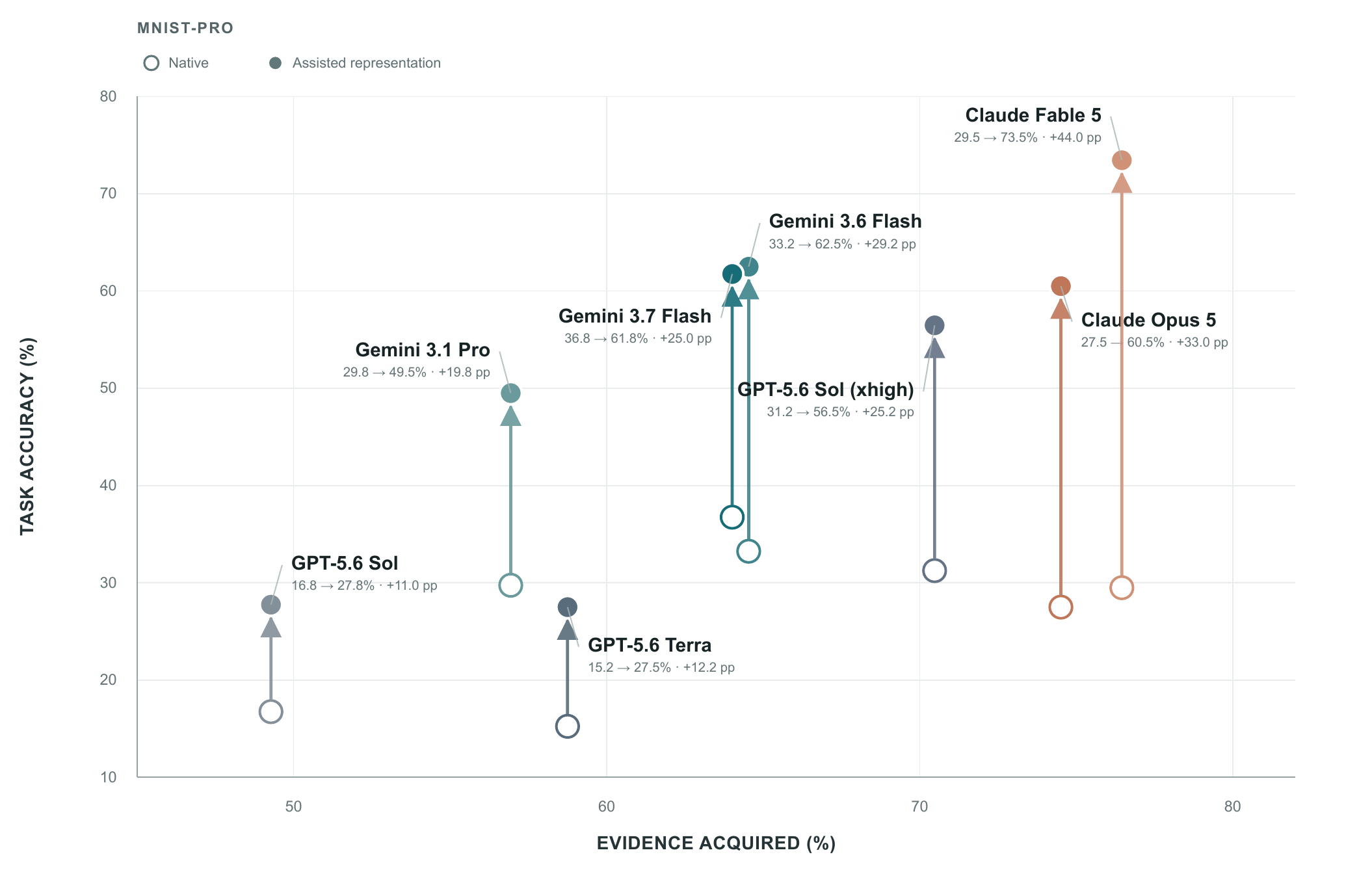}
        \caption{Native versus assisted-representation accuracy.}
        \label{fig:evidence-rescue}
    \end{subfigure}

    \caption{
        Acquiring visual evidence is not the same as effectively using it. Scores are equally averaged over Level 1 and Level 2 and over Image Only and Textual State.
    }
    \label{fig:evidence-to-decision}
\end{figure*}

Our analyses reveal two primary findings. First, acquiring visual evidence does not necessarily translate into the effective use of that evidence for improving accuracy (\Cref{fig:evidence-native}). Here, evidence acquisition is measured by the coverage of digit strokes observed by the models during exploration. The bottleneck may therefore lie in effectively composing the acquired observations into a useful perceptual-state representation and subsequently interpreting that state. Models such as \claudeopus{}, \claudefable{}, and \gptsolxhigh{} exhibit this pattern. On the other hand, \gptsol{} and \gptterra{} acquire substantially less evidence, which limits their ability to achieve high accuracy. The Gemini-3 models show a more balanced pattern: although their coverage is lower than that of \claudefable{}, \claudeopus{}, and \gptsolxhigh{}, they achieve higher accuracy, suggesting that they either use the acquired evidence more effectively or their acquired evidence is more informative.

Second, when we assist the agents by constructing a visual canvas that integrates their already observed glimpses (\Cref{fig:evidence-rescue}), a different pattern emerges. \claudeopus{} and \claudefable{} improve substantially and outperform the Gemini models under this assisted representation setting. \gptsolxhigh{} also improves, but still performs worse than \geminiseven{} and \geminisix{} despite acquiring more evidence. We hypothesize that this may be due to differences in the informativeness of the acquired evidence; for example, an agent may achieve high aggregate coverage by repeatedly observing nearby regions rather than exploring the most discriminative parts of the digit. In this case, only 38\% of \gptsolxhigh{} trajectories observe strokes from both digits, versus 76\% for \geminiseven{}. 

To disentangle evidence acquisition, perceptual-state construction, and interpretation of the constructed state, we perform an Offline Canvas experiment in which canvases are constructed from the trajectories of \gptsolxhigh{}, while the final predictor is replaced with \geminiseven{}. Because the trajectory and the resulting canvas remain fixed, any improvement obtained by changing only the final predictor directly reflects a difference in the ability to interpret the same constructed perceptual state. On Level 1, accuracy improves from 66.5\% with \gptsolxhigh{} as the predictor to 84.3\% with \geminiseven{}, a gain of 17.8 percentage points. Thus, even when provided with exactly the same acquired evidence and canvas representation, \geminiseven{} extracts substantially more useful information from it, exposing a clear interpretation bottleneck in \gptsolxhigh{}.

We observe a similar effect on Level 2 for Image Only trajectories, where replacing the \gptsolxhigh{} predictor with \geminiseven{} improves accuracy from 59\% to 74\%. However, the improvement is much smaller for canvases constructed from Textual State trajectories, increasing only from 26\% to 33\%. This difference reveals an additional evidence-acquisition bottleneck. As discussed previously, the Textual State trajectories of \gptsolxhigh{} observe both digits in only 38\% of episodes. When both digits are present in the resulting canvas, \geminiseven{} achieves 86.8\% accuracy, whereas its accuracy drops to 0\% when one of the digits is absent. Therefore, improving the final interpretation cannot recover information that was never acquired during exploration.

These results also show that aggregate coverage alone does not fully characterize the usefulness of the acquired evidence. On Level 2, \geminiseven{} achieves only 41\% accuracy when predicting from canvases constructed from its own Image Only trajectories, but reaches 74\% when predicting from \gptsolxhigh{} Image Only canvases. Thus, although \geminiseven{} is the stronger interpreter, the \gptsolxhigh{} trajectories provide a more informative perceptual state in this setting. The opposite pattern appears for Textual State trajectories: \geminiseven{} achieves 51\% on canvases constructed from its own trajectories but only 33\% on those constructed from \gptsolxhigh{} trajectories. Together, these controlled predictor-swap experiments separate three distinct factors in agentic perception: whether informative evidence is acquired, whether that evidence is organized into a usable perceptual-state representation, and whether the model can correctly interpret the resulting state.

We also observe that several agents exhibit premature decision-making, committing to a prediction despite having acquired limited visual evidence and achieving low coverage. This behavior is particularly visible for \glm{}, \gptsol{}, and \qwen{}. Interestingly, these agents use only a small fraction of the available visual sensing budget, even though the Level~1 budget is sufficient for a systematic sweep of the entire canvas. This suggests that the agents are often overly optimistic about the sufficiency of their current perceptual evidence and commit before resolving the remaining uncertainty. We hypothesize that this behavior may partly arise from the strong prior knowledge of MNIST digit shapes encoded in these models: a familiar partial stroke can support an early object hypothesis even when the observed evidence is insufficient to reliably identify the complete digit. This interpretation is consistent with prior findings that LLMs and VLMs can be overconfident under uncertainty and may answer even when additional visual observations are needed \citep{groot2024overconfidencekeyverbalizeduncertainty,zhang2026seeingisntknowingvlms}. A possible additional factor is confidence miscalibration introduced during post-training; prior work has associated RLHF with increased verbalized overconfidence \citep{leng2025tamingoverconfidencellmsreward}, although our experiments do not isolate the contribution of post-training.

Interestingly, \cite{toh2026scrambletoolbenchagentssearchexhaustively} observe the opposite behavior in ScrambleToolBench. There, agents operate in an unfamiliar environment in which tool semantics are deliberately obfuscated and can subsequently drift. Rather than efficiently inferring the underlying structural change, agents often fall back to exhaustive search instead of using deductive recovery strategies such as cycle tracing. We hypothesize that this contrast may arise partly from differences in task priors: \method{} presents a highly familiar perceptual domain that can induce early confidence, whereas ScrambleToolBench deliberately removes semantic priors and requires the latent environment structure to be discovered through interaction. More broadly, recent work on agentic search has also identified premature stopping and evidence sufficiency as central challenges in deciding how long an agent should continue gathering information \citep{kausik2026contextgatheringdecisionprocess,choubey2026dontstopearlyscalable}.

We characterize this \method{} behavior as \emph{overoptimistic stopping}. By committing while substantial sensing capacity remains, the agent behaves as though its current evidence is sufficient for a reliable decision. This confidence is often unwarranted: the agent terminates with low coverage and produces an incorrect prediction despite having enough remaining budget to systematically observe the entire Level~1 environment. Exhaustive sensing could provide complete visual evidence, although, as our other experiments show, complete evidence does not necessarily guarantee correct perceptual-state construction or interpretation.

\section{Related Work}

\paragraph{Vision-Language Benchmarks.}
Most vision-language benchmarks evaluate passive reasoning on fully observed, static scenes.
Standard benchmarks, such as MMBench \citep{liu2024mmbenchmultimodalmodelallaround}, MMMU \citep{yue2024mmmumassivemultidisciplinemultimodal}, MM-Vet \citep{yu2024mmvetevaluatinglargemultimodal}, SEED-Bench \citep{li2023seedbenchbenchmarkingmultimodalllms}, BLINK \citep{fu2024blinkmultimodallargelanguage}, and multitask evaluations like MMT-Bench \citep{ying2024mmtbenchcomprehensivemultimodalbenchmark}, present the visual input once to measure capabilities like visual recognition and commonsense reasoning.
Because these passive settings provide complete visual evidence at once, they do not evaluate the challenge of sequential information gain in real-world scenarios.
\method{} instead evaluates active, glimpse-based perception by requiring agents to sequentially acquire and integrate observations under a strict budget.

\paragraph{Active Perception and Visual Search.}
The active perception framework evaluates how agents control their gaze to resolve visual ambiguities, building upon early works on active vision \citep{bajcsy2016revisitingactiveperception}.
Active perception benchmarks for multimodal LLMs, such as ActiView \citep{wang2025activiewevaluatingactiveperception} and ActiveVision \citep{zhang2026examactiveobservers}, restrict the initial field of view and evaluate how models shift their gaze.
$V^*$ \citep{wu2023vguidedvisualsearch} uses LLM-guided visual search and a visual working memory to zoom into small, high-resolution details.
A limitation of these benchmarks is that they typically assume agents can maintain an expanding, unconstrained limit of all previous glimpses in their context window, without evaluating the challenge of state representation.
\method{} introduces controlled visual history limits ($\hist$) to evaluate agents under strict lookback constraint ($\hist=1$).
This allows us to evaluate how agents construct a perceptual state without relying on visual history accumulation, and to systematically test online and offline programmatic consolidation of past glimpses onto a merged visual canvas.

\paragraph{Working Memory under Partial Observability.}
To make optimal decisions under partial observability, agents must maintain a persistent representation of the environment across sequential steps.
In multimodal settings, memory evaluations are often framed within embodied navigation, such as ALFRED \citep{shridhar2020alfredbenchmarkinterpretinggrounded}, or focus on spatial-temporal mapping like SpatialBench \citep{xu2026spatialbenchbenchmarkingmultimodallarge} and VSI-Bench \citep{yang2025thinkingspacemultimodallarge}.
Other benchmarks, such as OSWorld \citep{xie2024osworldbenchmarkingmultimodalagents}, formalize GUI interactions as POMDPs, but require complex multi-step execution environments and specialized tool-use.
In these settings, overall performance is sensitive to physical noise, complex environment rendering, and tool-use proficiency, making it difficult to isolate the agent's state estimation capability.
\method{} abstracts away these variables to isolate and systematically evaluate agent performance under controlled, structured memory representations.

\section{Conclusion}
In this work, we introduced \method{}, a benchmark evaluating how multimodal AI agents explore and represent partially observed visual environments.
By transforming MNIST digit recognition into a sequential, glimpse-based search with lookback constraints, we analyzed the distinct roles of active exploration and state construction.
Our experiments revealed a clear performance gap between full-observability visual recognition and agentic perception.
We demonstrated that simply acquiring visual evidence does not guarantee accurate decision-making.
Instead, success depends on how agents represent and interpret their visual history.
While programmatically consolidating glimpses into a visual canvas improves accuracy for some models, others fail to interpret these canvases or to explore the most informative regions.
Our results show that building capable AI agents requires both active sensory control and structured state representation, rather than relying solely on expanding raw context windows.

\section*{Acknowledgment}
We wholeheartedly thank Deepmind and Google AI for awarding us with the GCP credits.

\bibliographystyle{plainnat}
\bibliography{main}


\appendix

\begin{figure}
    \centering
    \includegraphics[width=0.65\linewidth]{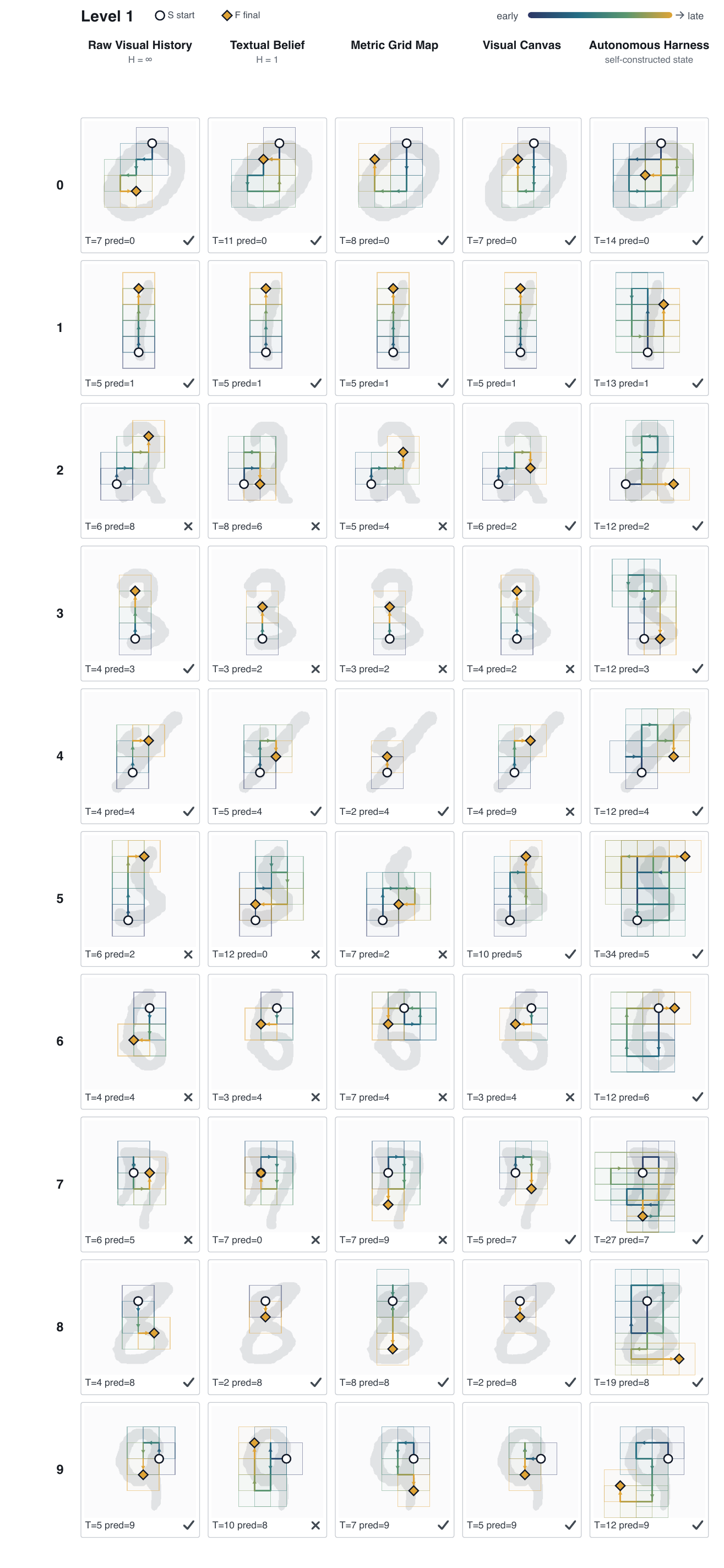}
    \caption{Exploration strategies by different state construction methods with \geminiseven{} for Level 1 experiments.}
    \label{fig:L1}
\end{figure}

\begin{figure}
    \centering
    \includegraphics[width=\linewidth]{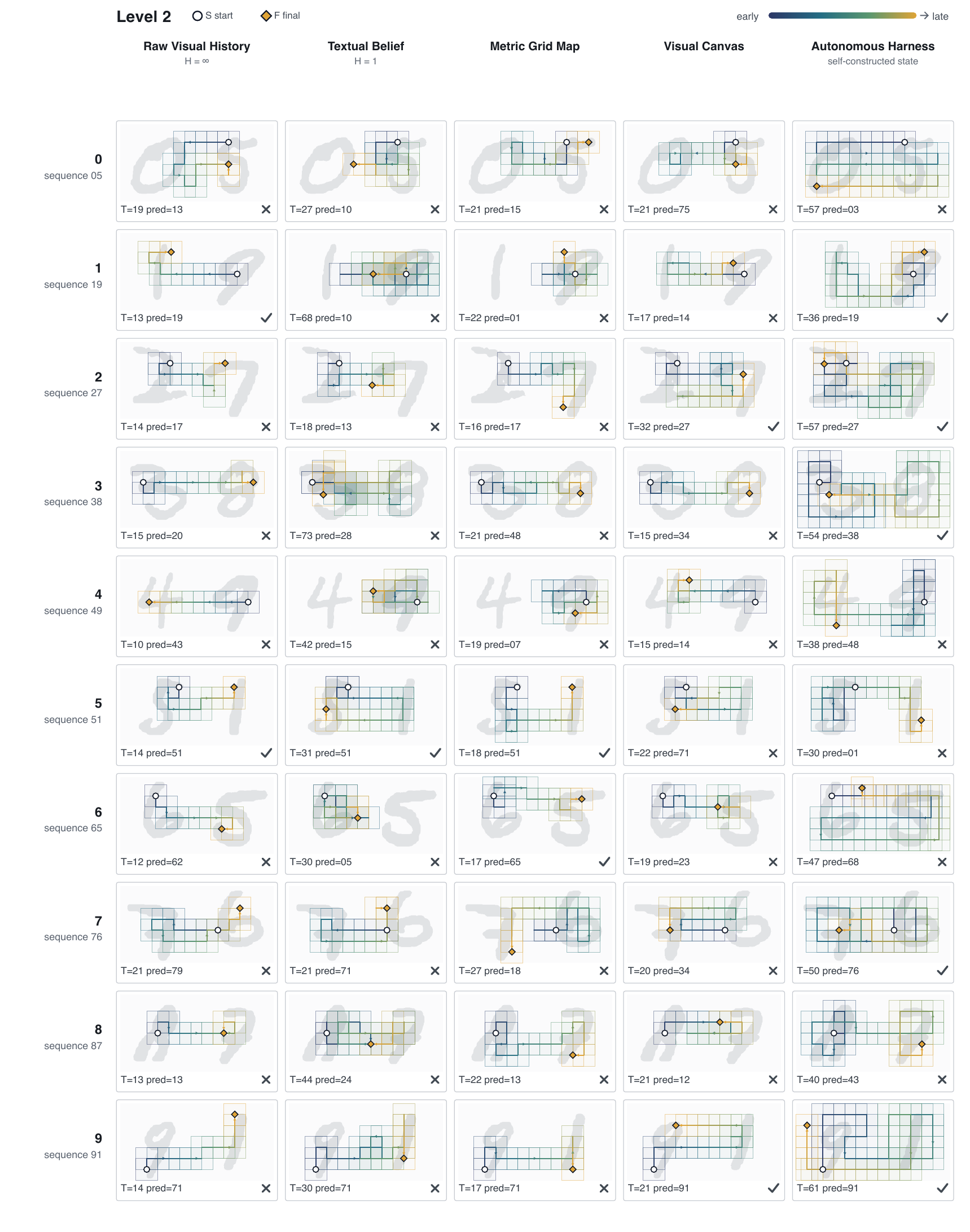}
    \caption{Exploration strategies by different state construction methods with \geminiseven{} for Level 2 experiments.}
    \label{fig:L2}
\end{figure}

\end{document}